\documentclass{clv3}
\usepackage{hyperref}
\usepackage{xcolor}
\definecolor{darkblue}{rgb}{0, 0, 0.5}
\hypersetup{colorlinks=true,citecolor=darkblue, linkcolor=darkblue, urlcolor=darkblue}

\usepackage{graphicx}
\usepackage{colortbl}
\usepackage{bbm}
\usepackage{amsfonts}
\usepackage[normalem]{ulem}
\useunder{\uline}{\ul}{}
\usepackage{amssymb}
\usepackage{framed}
\usepackage{color}
\usepackage{multicol}
\usepackage{multirow}
\usepackage{makecell}
\usepackage{amsmath}
\usepackage{booktabs}
\usepackage{algorithm}
\usepackage{wrapfig}
\usepackage{algpseudocode}
\usepackage{longtable}
\usepackage[]{tcolorbox}
\tcbuselibrary{skins}
\usepackage{enumitem}
\usepackage{bm}
\usepackage{silence}
\usepackage{anyfontsize}

\usepackage{arydshln}
\usepackage{fontawesome}

\begin{document}
\issue{xx}{yy}{2024}

\dochead{Short Paper}

\runningtitle{Alignment Self-Evaluation}

\runningauthor{Ye and Ng}


\title{Self-Judge: Selective Instruction Following with Alignment Self-Evaluation}

\author{Hai Ye}
\affil{Department of Computer Science \\ National University of Singapore\\\url{yehai@comp.nus.edu.sg}}

\author{Hwee Tou Ng}
\affil{Department of Computer Science \\ National University of Singapore\\\url{nght@comp.nus.edu.sg}}

\maketitle

\begin{abstract}
Pre-trained large language models (LLMs) can be tailored to adhere to human instructions through instruction tuning. However, due to shifts in the distribution of test-time data, they may not always execute instructions accurately, potentially generating factual errors or misaligned content when acting as chat assistants. To enhance the reliability of LLMs in following instructions, we propose the study of selective instruction following, whereby the system declines to execute instructions if the anticipated response quality is low. We train judge models that can predict numerical quality scores for model responses. 
To address data scarcity, we introduce \textsc{Self-J}, a novel self-training framework for developing judge models without needing human-annotated quality scores. Our method leverages the model's inherent self-evaluation capability to extract information about response quality from labeled instruction-tuning data. It incorporates a gold reference answer to facilitate self-evaluation and recalibrates by assessing the semantic similarity between the response sample and the gold reference. During the training phase, we implement self-distillation as a regularization technique to enhance the capability of reference-free estimation. 
To validate alignment evaluation on general instruction-following tasks, we collect large-scale high-quality instructions from Hugging Face for model training and evaluation. 
Extensive experiments on five open-source models show that our method correlates much more with GPT-4 than strong baselines, e.g., supervised models distilled from GPT-4 and GPT-3.5-turbo. Our analysis shows our model's strong generalization across domains. Additionally, our judge models serve as good reward models, e.g., boosting WizardLM-13B-V1.2 from 89.17 to 92.48 and from 12.03 to 15.90 in version v1 and v2 of AlpacaEval respectively using best-of-32 sampling with our judge models. 
Ranking 95 models from AlpacaEval, our judges show a high Kendall's $\tau$ correlation coefficient (0.63) with GPT-4. Our work underscores the potential of alignment self-evaluation in large language models\footnote{We release the source code and data used in this paper in \url{https://github.com/nusnlp/Self-J}}.

\end{abstract}

\section{Introduction}
By building generative transformers with larger number of model parameters and training on larger pre-training corpora, we can effectively create powerful large language models~(LLMs)~\cite{radford2019language,brown2020language}. 
LLMs have demonstrated strong generalization capabilities, enabling them to generalize to any task through in-context learning~\cite{brown2020language}. 
To make large language models more user-friendly, it is essential to further align them with human preferences, ensuring they generate content that is beneficial, safe, and follows user intentions~\cite{DBLP:conf/nips/Ouyang0JAWMZASR22}. 
Instruction tuning is an effective technique to align LLMs to human preferences. The models are fine-tuned to follow instructions described in natural language. Aiming to have a good generalization ability and to solve tasks across domains, instruction tuning usually involves diverse tasks. 
Instruction-following models have evolved into highly effective chat assistants for daily tasks, fulfilling various functions across different fields, e.g., content creation and customer support~\cite{ChatGPT}. 

Due to shifts in test-time distribution, models refined through instruction tuning may still produce outputs that are not aligned with human preferences, such as hallucinated content, unhelpful material, and irrelevant responses. To develop LLMs that adhere to human instructions, we propose studying selective instruction following where the system can decline to execute an instruction when it finds the response to be of low quality. We achieve this goal by exploring alignment evaluation that quantifies how well a model's output adheres to human preference. 
Alignment evaluation aims to measure the quality of model outputs on aspects such as helpfulness, correctness, relevance, etc~\cite{zheng2023judging}. This process is challenging since it involves diverse tasks and the model's outputs are expressed in natural language. 
The advanced capabilities of leading models like GPT-4 Turbo have increasingly been used to assess the performance of less powerful models. Multiple studies have shown a strong correlation between evaluations by these models and those conducted by humans~\cite{zheng2023judging,DBLP:journals/corr/abs-2305-14387,alpaca_eval}. Crowd-sourced human evaluation also plays a significant role. As seen through platforms like Chatbot Arena~\cite{zheng2023judging}, users provide preference feedback for two compared models, which is then used to calculate Elo ratings for gauging model performance. However, collecting human feedback is much slower and more costly. 
\begin{wrapfigure}[15]{r}{0.35\textwidth}
\setlength{\abovecaptionskip}{-0.3cm}
\vspace{-3mm}
 \centering
\includegraphics[width=\linewidth]{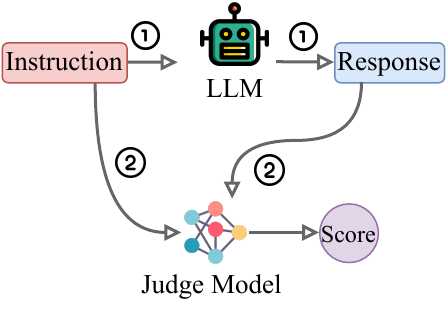} 
\caption{Selective instruction following with alignment evaluation. We train a judge model to rate an LLM's response with a numerical score.}
\label{fig:setup}
\end{wrapfigure}

In this work, we train judge models that can predict numerical quality scores for model outputs as a measure of alignment (see Fig.~\ref{fig:setup}). Independent of input from other LLMs or human annotations, we propose a \emph{self-training}-based method named \textsc{Self-J} by utilizing the model's self-evaluation capabilities. It aims to extract quality scores from labeled instruction-tuning data, and subsequently train the judge model using the generated scores. 
Alignment evaluation does not have a definitive metric (e.g., semantic similarity) that can automatically calculate quality scores by comparing the model response with a gold standard reference~\cite{zheng2023judging}. 
We rely on the instruction-tuned model itself to infer the quality scores for the model's own generated responses. To derive the score more accurately, we introduce the reference answer from the labeled instruction-tuning set to facilitate the model's self-evaluation. The reference answer offers a valuable supervision signal that facilitates self-evaluation, enabling the model to simply compare the sampled answer with the reference. To further reduce the noise in self-evaluation ratings, these ratings are recalibrated based on the semantic similarity between the model-generated samples and the gold-standard reference. This integration of semantic understanding and similarity metrics enables a more precise inference of quality scores. During the training phase, we introduce a self-distillation technique to regularize the training of judge models. Here, a teacher model, enhanced with additional reference answers, directs the training of the student model. It can enhance the estimation of reference-free quality during tests where reference answers are unavailable.

We evaluate \textsc{Self-J} on open-source models, e.g., Llama-2-Chat~\cite{touvron2023llama}. To validate alignment evaluation on general instruction-following tasks, such as coding, writing, etc, we collect a large number of high-quality instructions that cover practical questions from Hugging Face\footnote{https://huggingface.co/}, which is used for instruction tuning, judge modeling, and evaluation. From the collected instructions, we randomly sample 30k instructions for judge model tuning and another 1k instructions for alignment evaluation. 
We further expand the test set by including AlpacaEval~\cite{alpaca_eval}, a widely recognized benchmark for instruction-following tasks. AlpacaEval is considered a cross-domain evaluation set, making it suitable for testing the adaptability of our models across different contexts. 
In our evaluation, we report the correlation between measures with GPT-4's evaluation. 
We further evaluate the generalization ability of trained judge models, by first probing domain transfer from a source model to tested target models. On AlpacaEval, we augment the model with best-of-$N$ sampling by using our judge model as the reward model. 

Our contributions in this paper can be summarized as follows:
\vspace{-2mm}
\begin{enumerate}
    \item We introduce \textsc{Self-J}, a novel self-training-based framework for judge model learning without human annotated scores. It is independent of GPT-4 without distilling the estimation scores.
    
    \item We conduct extensive experiments with five open-source models for evaluation, which are Vicuna-13b, WizardLM-13b, Llama-2-chat-13b, Llama-2-chat-70b, and our instruction-tuned model using 87k of our collected instructions. The experimental results validate the effectiveness of our approach.
    
    \item \textsc{Self-J} surpasses GPT-3.5-turbo and GPT-4 distilled models on reference-free evaluation, and matches GPT-3.5-turbo on reference-based estimation.
    
    \item Serving as a reward model, \textsc{Self-J} empowers WizardLM-13B-V1.2 with best-of-32 sampling on AlpacaEval, by improving the performance from 89.17 to 92.48 and from 12.03 to 15.90 on v1 and v2 of AlpacaEval respectively. It even outperforms GPT-4-0613 on v2 evaluation.

    \item The tuned judge models achieve a high system-level Kendall's $\tau$ correlation coefficient (0.63) with GPT-4 for ranking 95 models submitted to AlpacaEval.
    
    \item A collection of high-quality instructions on a large scale has been compiled for analysis. By fine-tuning Llama-2-13b with randomly chosen 87k instructions and GPT-3.5 Turbo responses, we can match Llama-2-13b-Chat's performance on AlpacaEval.
\end{enumerate}
\vspace{-2mm}

\section{Related Work}
\subsection{Instruction Tuning}
Aligning large language models to human preference is important since it stops the generation of harmful and useless content. Reinforcement learning from human feedback~(RLHF) has become a standard approach to align LLMs. This involves initial instruction fine-tuning followed by reinforcement learning, as detailed by \cite{DBLP:conf/nips/Ouyang0JAWMZASR22}. The focus has recently shifted towards self-alignment strategies, which aim to align LLMs more efficiently and cost-effectively. Key areas of exploration include instruction generation and reward modeling. \citet{DBLP:conf/acl/WangKMLSKH23} demonstrate the potential of generating instructions and responses using in-context learning with just 175 human-labeled examples. Following this, research by \cite{DBLP:journals/corr/abs-2305-03047} has shown that model alignment can be achieved with as few as five labeled examples. Another study from \cite{DBLP:journals/corr/abs-2308-06259} introduces the concept of back-translating responses into instructions as a novel alignment technique. Furthermore, \citet{DBLP:journals/corr/abs-2305-11206} highlight the significance of instruction diversity over quantity for effective alignment. For reward modeling, recent works have leveraged large language models to provide feedback. A study by \cite{DBLP:journals/corr/abs-2310-05910} utilizes an LLM to offer preference feedback on model responses based on a set of principles. \citet{DBLP:journals/corr/abs-2212-08073} focus on generating non-harmful prompts and employ LLMs to provide feedback, aiming to identify less harmful responses.

\subsection{Alignment Evaluation}
There has been a growing interest in the automatic evaluation of chat assistant performance. This trend is driven by the absence of reliable evaluation metrics and the prohibitive costs of human evaluation. Recent studies have leveraged state-of-the-art models, such as GPT-4, to assess the capabilities of less sophisticated models~\cite{zheng2023judging,alpaca_eval}. These investigations have uncovered a strong correlation between evaluations conducted by GPT-4 and those performed by humans~\cite{zheng2023judging,alpaca_eval,DBLP:journals/corr/abs-2305-14387}. Nevertheless, the proprietary nature and the substantial expense linked to the use of GPT-4 have spurred initiatives aimed at creating open-source judge models. These models are designed to offer consistent and cost-effective evaluations~\cite{li2023generative,wang2023shepherd,wang2023pandalm}.
\citet{li2023generative} develop an open-source model, auto-j, which is distilled from GPT-4. This model is designed to assess alignments by furnishing both a critique and a score. Similarly, \citet{wang2023pandalm} introduce a model also distilled from GPT-4, designed to express a preference between pairs of responses to a given question, which is particularly useful for optimizing hyper-parameters during instruction tuning. 
\citet{cui2023ultrafeedback} focus on constructing an open-source reward model that leverages preference feedback data from GPT-4 for reward modeling. This model is intended to support the community in developing more effective alignment algorithms. 
Another contribution by \cite{wang2023shepherd} involves the creation of a model capable of generating critiques for model responses, which is achieved by aggregating critique data from the Internet. To the best of our knowledge, our method represents the \emph{first} attempt to train a judge model that does not depend on demonstration scores from GPT-4 for its training.

\subsection{AI Feedback}
Reinforcement learning from human feedback (RLHF) has traditionally been a resource-intensive process, which needs significant human effort to collect feedback. Recent work has introduced a new approach, reinforcement learning from AI feedback (RLAIF), which leverages large language models to provide feedback, thereby replacing the need for human input. \citet{DBLP:journals/corr/abs-2212-08073} utilize LLMs to offer preference feedback on the outcomes of harmful prompts. This feedback is then used to train a reward model aimed at aligning with harmless content. \citet{yuan2024self} prompt LLMs to assign numerical scores to model responses. These scored responses are subsequently paired and used in iterative training of the model, employing direct preference optimization~\cite{DBLP:journals/corr/abs-2305-18290}. Another study investigates self-alignment by converting responses back into instructions, with LLMs filtering out instructions of lower quality~\cite{DBLP:journals/corr/abs-2308-06259}. \citet{lee2023rlaif} delve into the utilization of AI feedback specifically for the task of summarization. Furthermore, the embedding of principles within reward modeling through feedback from large language models has been proposed as a novel strategy for enhancing alignment~\cite{DBLP:journals/corr/abs-2310-05910}. Our research investigates self-alignment evaluation, which is closely linked to the concept of AI feedback.

\subsection{Open-Source Instructions}
The pursuit of open-source advancements within the community extends beyond just models to include data as well. \citet{DBLP:conf/acl/WangKMLSKH23} and \citet{alpaca} have leveraged in-context learning to generate a large set of instructions. Similarly, \citet{DatabricksBlog2023DollyV2} have opted for a human-centric approach, gathering instructions and responses to compile an open-source dataset featuring 15k instructions. \citet{DBLP:conf/icml/LongpreHVWCTZLZ23} have embarked on a mission to amass a large collection of instructions with a focus on traditional NLP tasks, such as natural language understanding and question answering. 
\citet{xu2023wizardlm} have utilized ChatGPT to generate challenging instructions, discovering that fine-tuning LLMs with difficult prompts can notably enhance model performance. Similarly, \citet{ding2023enhancing} have employed ChatGPT for generating a vast array of dialogue data, further illustrating the versatility of LLMs in simulating complex conversational scenarios. 
The instruction set of ShareGPT has been widely used for instruction-tuning~\cite{vicuna2023}. In our work, we compile a comprehensive collection of practical and high-quality instructions from Hugging Face, contributing to the pool of resources available for enhancing the effectiveness and efficiency of instruction-tuning LLMs.

\subsection{Uncertainty Estimation for Language Generation}
There is a growing interest in measuring uncertainty in conditional language generation. \citet{ren2022out} introduce an approach that involves training a lightweight model, incorporating both encoder and decoder embeddings derived from transformers. This model is tailored for selective instruction following as well as identification of out-of-distribution (OOD) samples. Crucially, their research highlights the inadequacy of perplexity as a reliable measure for gauging a model's confidence level in generated text, suggesting the need for alternative metrics. 
Building on this quest for better uncertainty metrics, \citet{kuhn2023semantic} have proposed the concept of semantic entropy. This method employs a natural language inference (NLI) model to estimate the semantic discrepancies among several model responses to an input, providing a more nuanced understanding of model uncertainty. 
Similarly, the notion of sampling variance has been explored by \cite{huang2023look}, which quantifies the semantic variability across multiple samples generated by a model. This technique offers another layer of insights into a model's performance by assessing the consistency of its outputs. 
In the field of machine translation (MT), efforts by \cite{guerreiro2022looking} and \cite{rei2020comet} have delved into the training of classifiers dedicated to the estimation of translation quality. The work of \cite{chollampatt-ng-2018-neural} and \cite{qorib-ng-2023} concerns quality estimation of grammatical error correction.

\begin{figure}[t]
\begin{tcolorbox}[
  enhanced,
  colback=white,
  colframe=black,
  coltext=black,
  boxsep=0pt,
  arc=0pt,
  outer arc=0pt,
  boxrule=0.8pt, 
]

Please act as an impartial judge and evaluate the quality of the response provided by an AI assistant to the user question displayed below. Your evaluation should consider factors such as the helpfulness, relevance, accuracy, depth, creativity, and level of detail of the response. You will be given a reference answer and the assistant's answer. Begin your evaluation by comparing the assistant's answer with the reference answer. Be as objective as possible. After providing your explanation, in the last new line, you must rate the response on a scale of 1 to 10 by strictly following this format: ``\{``rating'': your rating\}''. For example, ``\{``rating'': 5\}''.

\vspace{\baselineskip}

[Question]

\{question\}

\vspace{\baselineskip}

[The Start of Reference Answer]

\{reference answer\}

\vspace{\baselineskip}

[The End of Reference Answer]

\vspace{\baselineskip}

[The Start of Assistant's Answer]

\{answer\}
\vspace{\baselineskip}

[The End of Assistant's Answer]

\end{tcolorbox}
\caption{The prompt used for alignment evaluation. It is used by GPT-4 and other open-source models studied in this work. We use the version of reference-based evaluation, where a reference answer is required for evaluation. The prompt is adopted from \citet{zheng2023judging} and \citet{DBLP:journals/corr/abs-2305-14314}. As indicated by \citet{zheng2023judging}, reference answers can improve the performance of GPT-4's evaluations on reasoning tasks, such as coding problems and mathematical questions.}
\label{tab:prompt-gpt-eval}
\end{figure}

\section{Preliminary}

\subsection{Instruction Fine-Tuning}
To align a large language model with instruction tuning, we need a labeled instruction set that involves diverse tasks, such as coding, logical reasoning, knowledge verification, strategic planning, creative ideation, etc. We denote the labeled set as $\mathcal{D}_{train} =\{\bm{x}, \bm{y}\}$, and the LLM is fine-tuned to generate the response $\bm{y}$ conditioned on the input instruction $\bm{x}$. After fine-tuning, a reinforcement learning stage can be applied to further align the model to human preference~\cite{DBLP:journals/corr/abs-2212-08073}. At inference time, the instruction-tuned LLM $\mathcal{F}$ samples a response $\bm{y}'$ conditioned on the instruction $\bm{x}$, i.e., $\bm{y}' \sim \mathcal{F}(\bm{y}' |\bm{x})$.

\subsection{Alignment Evaluation} 
\noindent{\textbf{Evaluation Criteria.}} \ For instruction-following LLMs, alignment evaluation aims to measure how well the model outputs align with human preference. Preference for desired outputs may vary among different users. In this work, we follow previous work that focuses on evaluating the preferences shared by most users which are helpfulness, relevance, accuracy, depth, creativity, and level of detail of the response~\cite{zheng2023judging}. As demonstrated by \citet{zheng2023judging}, when prompting GPT-4 as the evaluator using these aspects of preference, it can have a high correlation with human evaluations. Below is a breakdown of each aspect:
\begin{itemize} 
     \item \textbf{Helpfulness} measures whether the model response effectively addresses the instruction. A helpful response should provide valuable solutions to solve the problems of the user. 
    
    \item \textbf{Relevance} indicates how closely the model response aligns with the user question or the topic of the question. A highly relevant response should directly address the user's problem without discussion about unrelated areas. 

    \item \textbf{Accuracy} evaluates the correctness of the information provided in the response. An accurate response should be free from errors and based on knowledge that can be traced to some sources. 

    \item \textbf{Depth} focuses on the thoroughness and comprehensiveness of the response. An answer with depth should go beyond a superficial response, offering a detailed understanding or insight.

    \item \textbf{Creativity} is about the novelty of the response. Creative responses should introduce new ideas, solutions, or perspectives that are not conventional. 

    \item \textbf{Level of detail} assesses the amount of information contained in the response. A detailed response should contain specific examples, data, or explanations to support the main arguments. 
\end{itemize}

By focusing on the six aspects during evaluation, a comprehensive score will be induced as a measure of the level of alignment. As shown in the prompt in Fig.~\ref{tab:prompt-gpt-eval}, the alignment score is an integer from 1 to 10. 

\noindent{\textbf{Types of Alignment Evaluation.}} \ By evaluating a response $\bm{y}' \sim \mathcal{F}(\bm{y}' | \bm{x})$, an estimator $\mathcal{J}_\theta$ parameterized by $\theta$ produces a quality score $z$. There are two types of alignment evaluation: reference-free and reference-based estimation.

$\triangleright$ \textbf{Reference-free:} It considers the scenario that the reference answer is usually not available at test time. The estimator takes in $\bm{x}$ and $\bm{y}'$ for evaluation, i.e., $z \sim \mathcal{J}_\theta(z | \bm{x}, \bm{y}')$. 

$\triangleright$ \textbf{Reference-based:} It is an oracle setting where the reference answer $\bm{y}$ is available, i.e., $z \sim \mathcal{J}_\theta(z | \bm{x}, \bm{y}', \bm{y})$. As pointed out by \citet{zheng2023judging}, LLMs are limited in their reasoning capability. They fall short in grading reasoning tasks since they are not aware of the correct answer to the question during evaluation. Providing a reference answer can enhance the ability of the LLMs to assess such questions. 

In this work, we are interested in reference-free estimation since reference answers are not available at test time. We also aim to benefit reference-free estimation from reference-based evaluation through self-distillation. 

\begin{figure*}[t]
\centering
\includegraphics[width=\textwidth]{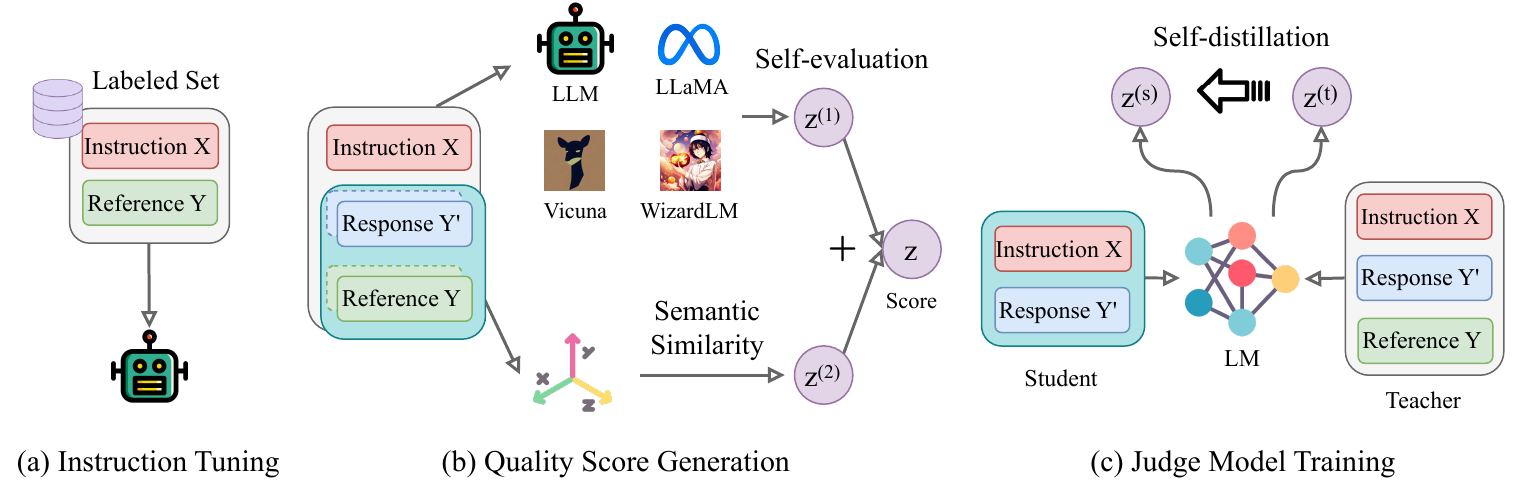} 
\caption{Illustration of \textsc{Self-J}~(see also the pseudocode of Algorithm \ref{code:self-j}). (a) We first conduct instruction tuning on a pre-trained LLM~(or directly using an existing instruction-tuned model, e.g., Vicuna). (b) We generate quality scores with model self-evaluation recalibrated by a semantic similarity score. (c) With the generated quality scores, we train a judge model through self-distillation.}
\label{fig:self-j}
\end{figure*}

\subsection{Selective Instruction Following}\label{sec:select-gen}
For selective instruction following, the system can refuse to execute the instruction (or display the answer) when the generated answer is of low quality. The alignment score represents the quality of a model's generated response.
We follow the setting of selective prediction~\cite{kamath-etal-2020-selective} to formulate selective instruction following. At test time, given an instruction $\bm{x}$ without reference answer, the quality score $z$ is generated conditioned on the instruction $\bm{x}$ and model response $\bm{y}'$, i.e., $z \sim \mathcal{J}_\theta(z | \bm{x}, \bm{y}')$. With a threshold $\eta \in  \mathbb{R}$, if $z \geq  \eta$, the response $\bm{y}'$ is accepted; otherwise, the response is discarded.

\section{Uncertainty Measurement}
Uncertainty represents the confidence level of a model's output. It has been widely studied and used for selective prediction~(or generation) and out-of-distribution~(OOD) detection~\cite{ren2022out}. 

\noindent{\textbf{PPL.}} \ For language generation, perplexity is a simple method to measure the uncertainty of a model, which is monotonically related to the average of negative log-likelihood over output tokens, i.e., $-\frac{1}{|\bm{y}'|}\sum_{t=1}^{|\bm{y}'|} \log p(y'_t|y'_{<t},\bm{x})$. However, recent work points out that it is not effective for selective language generation tasks~\cite{ren2022out}. 

\noindent{\textbf{Sampling Variance}.} \ A better way to estimate the uncertainty of language generation is the variance ratio for the original output~(VRO) with extra sampled responses~\cite{huang2023look}. 
Except for the original response $\bm{y}'$, we further sample extra $K$ responses $\{\bm{y}'_k\}_{k=1}^K$. The variance over the sampled responses is calculated as follows:
\begin{equation}
    -\frac{1}{K} \sum_{k=1}^K S(\bm{y}', \bm{y}'_k)
    \label{eq:vro}
\end{equation}
where $S(\cdot, \cdot)$ measures the semantic similarity between the original prediction and another sampled response. More similar responses will have a smaller variance, and the model is expected to be more certain about its generation. We calculate cosine similarity over the response embeddings.

\section{Method}
Here, we introduce our method \textsc{Self-J} for judge model training, which aims to utilize self-training to train judge models without human annotations. 
Fig.~\ref{fig:self-j} is an overview of our method. Briefly speaking, \textsc{Self-J} begins by instruction-tuning an LLM~(or using an existing instruction-tuned LLM). It then self-generates quality scores for model responses~($\S$\ref{sec:data-gen}), using these to fine-tune a judge model through self-distillation~($\S$\ref{sec:self-distil}). 

Tasks in instruction tuning, unlike more established natural language generation (NLG) tasks such as machine translation, summarization, or grammatical error correction, lack a standard evaluation metric, which complicates automatic quality scoring using metrics like BLEU~\cite{papineni2002bleu}, Rouge~\cite{lin2004rouge}, or $M^2$~\cite{dahlmeier2012}. We therefore present a more effective way to assess the quality scores. On the labeled instruction set $\mathcal{D}_{train} = \{\bm{x}, \bm{y}\}$, we first sample model responses $\bm{y}'$ from an instruction-tuned LLM $\mathcal{F}$, i.e., $\bm{y}' \sim \mathcal{F}(\bm{y}' | \bm{x})$. After sampling, we obtain the training set $\mathcal{D}'_{train}=\{\bm{x}, \bm{y}, \bm{y}'\}$ with model responses.

Since there are no labeled quality scores, we aim to train the judge model with self-training, where we first generate quality scores and then train the judge model with the generated scores. 
Our goal is to tune an LLM to generate an accurate quality score $z \sim \mathcal{J}_\theta(\cdot | \bm{x}, \bm{y}')$ given $(\bm{x},\bm{y}') \sim p(\bm{x}, \bm{y}')$, where the judge model $\mathcal{J}_\theta$ is parameterized by $\theta$. Suppose we have the true distribution $q(z|\bm{x}, \bm{y}')$ that generates the quality score, we can tune the judge model $\mathcal{J}_\theta$ with cross entropy minimization:
\begin{equation}
    \mathcal{L}_\theta(\bm{x},\bm{y}') = - \sum_{z} q(z|\bm{x}, \bm{y}')\log \mathcal{J}_\theta(z|\bm{x},\bm{y}') 
\end{equation}
which aims to minimize the difference between the two distributions $q(z|\bm{x}, \bm{y}')$ and $\mathcal{J}_\theta(z|\bm{x},\bm{y}')$.

We can approximate the true distribution with the instruction-tuned model $\mathcal{F}$ where we can prompt the model $\mathcal{F}$ to generate the quality score $z \sim \mathcal{F}(\cdot | \bm{x}, \bm{y}')$ for the data $(\bm{x}, \bm{y}')$. However, the instruction-tuned model may not be good at estimating the quality of self-generated responses, especially for questions that need strong reasoning abilities~\cite{zheng2023judging}. As pointed out by \citet{zheng2023judging} and demonstrated by our experiments, providing a reference answer to the model can enhance the model's performance on response quality estimation. To better approximate the true distribution, we further introduce the reference answers $\bm{y}_i$, and $q(z|\bm{x},\bm{y}')$ can be further derived as follows:
\begin{equation}
\begin{aligned}
    q(z|\bm{x},\bm{y}') 
    & = \sum_{\bm{y}_i} q(z, \bm{y}_i | \bm{x}, \bm{y}') \\
    & = \sum_{\bm{y}_i} q(z|\bm{x},\bm{y}_i, \bm{y}') q(\bm{y}_i | \bm{x}, \bm{y}') \\ 
    & = \sum_{\bm{y}_i} q(z|\bm{x},\bm{y}_i, \bm{y}') \underbrace{\frac{q(\bm{y}_i, \bm{y}'|\bm{x})}{q(\bm{y}'|\bm{x})}}_{w_i}
\label{eq:proved_eq}
\end{aligned}
\end{equation}
In the given equation, the distribution of $q(z|\bm{x},\bm{y}')$ represents the weighted sum of $q(z|\bm{x},\bm{y}_i, \bm{y}')$ across different reference answers. Here, the weight $w_i$ is $\frac{q(\bm{y}_i, \bm{y}'|\bm{x})}{q(\bm{y}'|\bm{x})}$. Since $q(\bm{y}'|\bm{x})$ is constant for varying references, it can be omitted, allowing us to use $q(\bm{y}_i, \bm{y}'|\bm{x})$ directly as the weight. Assuming $z_i \sim q(\cdot|\bm{x}, \bm{y}_i, \bm{y}')$, the aggregated score $z$ is computed as $\sum_i w_i \cdot z_i$. We hypothesize that all reference answers \(\bm{y}_i\) are of comparable quality and exert a uniform influence on quality assessment. This assumption enables us to reduce the number of reference answers required when applying Eq.~\ref{eq:proved_eq} to compute quality scores.

\subsection{Quality Score Generation}\label{sec:data-gen}
In this section, we generate the quality scores for training based on Eq.~\ref{eq:proved_eq}, utilizing the training dataset \( \mathcal{D}'_{train} = \{\bm{x}, \bm{y}, \bm{y}'\} \). According to Eq.~\ref{eq:proved_eq}, the score \( z \) is a recalibrated score of \( q(z|\bm{x}, \bm{y}, \bm{y}') \) by using \( q(\bm{y}, \bm{y}'|\bm{x}) \). This process ensures that both factors are considered when determining the quality of the scores.

We employ the instruction-tuned model \( \mathcal{F} \) to initialize \( q(z|\bm{x},\bm{y},\bm{y}') \), capitalizing on the robust generalization capabilities of LLMs for self-assessment. By referencing the prompt illustrated in Fig.~\ref{tab:prompt-gpt-eval}, we input the tuple \( ( \bm{x}, \bm{y}', \bm{y} ) \) into the instruction-tuned model $\mathcal{F}$. This setup prompts the model to generate an integer quality score \( z \) for the response, where \( z \) ranges from 1 to 10~($1 \leq z \leq 10$).

In our framework, \( q(\bm{y}, \bm{y}' | \bm{x}) \) reflects the semantic similarity between the model's response \( \bm{y}' \) and the gold standard reference \( \bm{y} \). A response that exhibits a high degree of semantic agreement with the gold reference typically indicates superior quality; on the other hand, lower similarity often points to reduced quality. When the self-evaluation score produced by the model \( \mathcal{F} \) proves to be inaccurate, leveraging the semantic similarity can serve to recalibrate the score, effectively acting as an ensemble mechanism. To assess this consistency, we utilize the cosine similarity between the embedded representations of the model response and the gold reference.

Given that both self-evaluation by the instruction-tuned model \( \mathcal{F} \) and calculation of cosine similarity are susceptible to noise, we derive the score \( z \) by taking the weighted average of the self-evaluation score \( z^{(1)} \) and the cosine similarity score \( z^{(2)} \)\footnote{However, other methods can also be applied to integrate these two scores such as multiplying the two values.}:
\begin{equation}
    z = \alpha z^{(1)} + (1 - \alpha) z^{(2)}
    \label{eq:score_combine}
\end{equation}
This approach helps to mitigate the impact of any inaccuracies that might arise from either individual metric. 
Since $z^{(1)}$ is an integer ranging from $1$ to $10$, and $z^{(2)}$ is a real number in the interval [-1,1], we discretize $z^{(2)}$ by evenly distributing its values across 1 to 10 so that $z^{(1)}$ and $z^{(2)}$ are on the same scale. After combination, over the training set, we further adjust the score $z$ to conform to a uniform distribution across the range from 1 to 10; $z$ is also an integer and $1 \leq z \leq 10$.

\noindent{\textbf{Search for optimal $\alpha^*$.}} \ 
To combine the two scores more effectively, we use a small development set $\mathcal{Q}_{dev} = \{\bm{x}, \bm{y}', \bm{y}, z^*\}$ ($150$ samples in our experiments) with human-labeled scores $z^*$ to find the optimal $\alpha^*$. 
To effectively combine the self-evaluation and cosine similarity scores, given the small size of the development set, we standardized each set of scores using Z-score normalization. Then we iteratively search for the optimal value of \(\alpha\) within the range from 0 to 1, using a step size of 0.1. For each candidate \(\alpha\), we compute the correlation between the scores of \(z\) and the labeled scores \(z^*\). The \(\alpha\) that yields the highest correlation is selected as the optimal value, denoted as \(\alpha^*\).

\begin{algorithm}[t]
\caption{Pseudocode for \textsc{Self-J}}
\label{code:self-j}
\begin{algorithmic}[1]
\State \textbf{Input:} Instruction set $\mathcal{D}_{train} = \{\bm{x}, \bm{y}\}$; Dev set $\mathcal{Q}_{dev} = \{\bm{x}, \bm{y}', \bm{y}, z^*\}$ with labeled quality scores; Instruction-tuned LLM $\mathcal{F}$.

\State Sample model responses with model $\mathcal{F}$ on $\mathcal{D}_{train}$:
\[
\mathcal{D}'_{train} = \Big \{ ( \bm{x}_i, \bm{y}'_i, \bm{y}_i ) \ | \ \bm{y}'_i \sim \mathcal{F}(\bm{y}' | \bm{x}_i), \forall  \bm{x}_i \in \mathcal{D}_{train} \Big \}.
\]

\State Generate self-evaluation ratings with model $\mathcal{F}$ on $\mathcal{D}'_{train}$: 
\[
\mathcal{Z}^{(1)} = \left \{ z^{(1)}_i \ | \ z^{(1)}_i \sim \mathcal{F}(z | \bm{x}_i, \bm{y}_i', \bm{y}_i), \forall ( \bm{x}_i, \bm{y}'_i, \bm{y}_i ) \in \mathcal{D}'_{train} \right \}.
\]

\State Calculate cosine similarities on $\mathcal{D}'_{train}$: 
\[
\mathcal{Z}^{(2)} = \left \{ z^{(2)} \ | \ z^{(2)} \sim S(\bm{y}'_i, \bm{y}_i), \forall ( \bm{y}'_i, \bm{y}_i ) \in \mathcal{D}'_{train} \right \}.
\]

\State Search for an optimal $\alpha^*$ on the dev set $\mathcal{Q}_{dev}$.

\State Combine self-evaluation ratings and cosine similarities: 
\[
\mathcal{Q}_{train} = \left \{ ( \bm{x}_i, \bm{y}'_i, \bm{y}_i, z_i ) \  | \ z_i = \alpha^* z^{(1)}_i + (1 - \alpha^*) z^{(2)}_i, \forall  z^{(1)}_i \in \mathcal{Z}^{(1)}, z^{(2)}_i \in \mathcal{Z}^{(2)} \right \}.
\]

\State Adjust scores $z$ to a uniform distribution; $z$ is an integer and $1 \leq z \leq 10$.

\State Train a language model on $\mathcal{Q}_{train}$ with self-distillation loss in Equation~\ref{eq:sd}.

\State \textbf{Output:} Judge model $\mathcal{J}_\theta$.

\end{algorithmic}
\end{algorithm}

\subsection{Self-Distillation for Judge Model Training}\label{sec:self-distil}
After creating a training set $\mathcal{Q}_{train} = \{\bm{x}, \bm{y}',\bm{y},z\}$ with quality scores of model responses, we proceed to fine-tune a pre-trained LLM to develop the judge model $\mathcal{J}_\theta$. For reference-free estimation, the judge model, taking the input instruction $\bm{x}$ and the model response $\bm{y}'$, is trained to predict a numerical score $z$, where $z \sim \mathcal{J}_\theta(z | \bm{x}, \bm{y}')$. Using the training set $\mathcal{Q}_{train}$, we fine-tune the LLM to minimize the negative log-likelihood:
\begin{equation}
    \mathcal{L}_{NLL}(\theta) = - \frac{1}{|\mathcal{Q}_{train}|} \sum_{i=1}^{|\mathcal{Q}_{train}|} \log  p(z_i|\bm{x}_i,\bm{y}'_i)
    \label{eq:nll}
\end{equation}
where we use the template shown in Fig.~\ref{fig:template-judge-without-refer} to include $\bm{x}$ and $\bm{y}'$ as the input context. For implementation convenience, we subtract 1 from the scores in the training set so that during training, the range of z is from 0 to 9.

During testing, the judge model estimates the quality score $z$ using only the context of the input instruction and model response, since usually, no reference response is available to the judge model. Ideally, if the reference answer is available, it would significantly aid the model in making more precise estimation of the score $z$. However, the reference answer is typically absent during testing. To address this problem, we introduce a self-distillation approach to train the judge model. It involves optimizing a \emph{teacher} objective that incorporates the reference answer for quality estimation, i.e., $p(z|\bm{x},\bm{y}',\bm{y})$, and then distilling the ability of the \emph{teacher} into a \emph{student} model that performs reference-free estimation, i.e., $p(z|\bm{x},\bm{y}')$. 

We optimize a KL-divergence loss for self-distillation. By considering the self-distillation objective, our final training loss is defined as:
\begin{equation}
\begin{aligned}
    \mathcal{L}_{SD}(\theta) = -\frac{1}{|\mathcal{Q}_{train}|} \sum_{i=1}^{|\mathcal{Q}_{train}|} \Big ( &  \log p(z_i|\bm{x}_i, \bm{y}'_i,\bm{y}_i) 
    \\ & + \beta \log p(z_i|\bm{x}_i,\bm{y}'_i) 
    \\ & - \gamma {\text{KL}}\left [ p(z_i|\bm{x}_i,\bm{y}'_i) \| p(z_i|\bm{x}_i, \bm{y}'_i, \bm{y}_i) \right ] \Big )
\end{aligned}
\label{eq:sd}
\end{equation}
Here, ``SD'' stands for ``self-distillation'', with $\beta$ and $\gamma$ serving as hyper-parameters. We fine-tune the same model using both teacher and student objectives, which is why this process is termed self-distillation. For each batch of training data, the model undergoes two forward passes—one for the teacher objective and one for the student objective. During the optimization of the KL loss, gradient back-propagation in the teacher is omitted. 
For the \emph{teacher} objective, we use the template in Fig.~\ref{fig:template-judge-with-refer} to include $\bm{x}$, $\bm{y}'$, and $\bm{y}$ as the input context. The pseudocode in Algorithm \ref{code:self-j} displays the detailed training procedure of \textsc{Self-J}.

After training, our model—referred to as the judge model—can perform both reference-free and reference-based evaluations. Focusing on the reference-free method, to determine the quality \( z \) of a response \( \bm{y}' \), we compute the expected score \( z \) as follows:
\begin{equation}
    z = \sum_{c_i = 0}^{C} c_i \cdot p(c_i|\bm{x}, \bm{y}')
\end{equation}
In this equation, \( c_i \) represents each possible score class, ranging from 0 to \( C \) (with \( C=9 \) in our case). The model predicts the probability \( p(c_i|\bm{x}, \bm{y}') \) for each score class. Thus, \( z \) is calculated as a weighted average, integrating the probabilities of each score class effectively.

\begin{figure*}[t]
    \centering
    \begin{minipage}{0.52\textwidth}
        \centering
        \includegraphics[width=\linewidth]{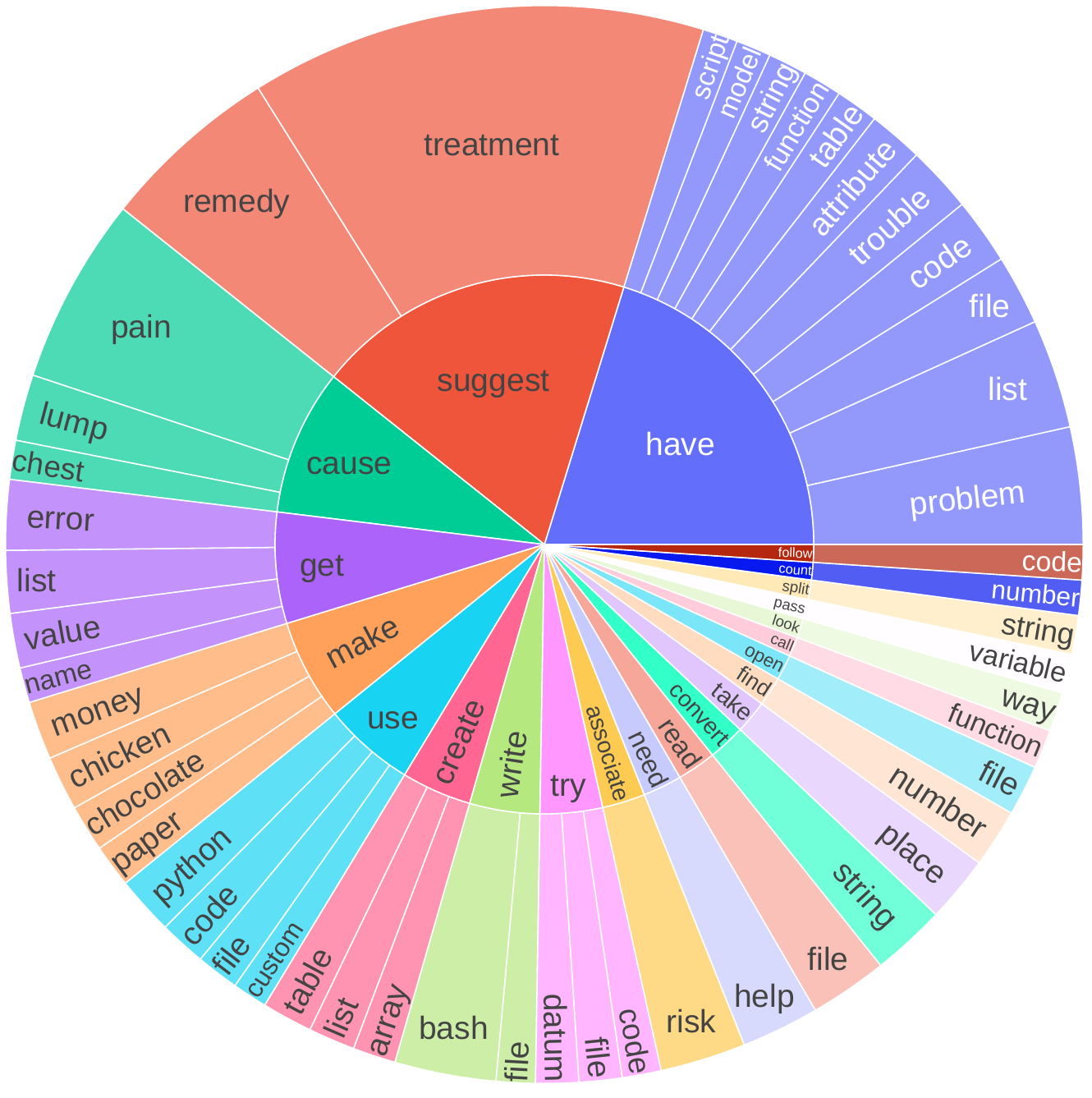} 
        \caption{Diversity of our instruction set~(with 30k random samples for judge model training), showing root verbs in the inner circle and their first nouns in the outer circle.}
        \label{fig:inst-diversity}
    \end{minipage}
    \hspace{1pt}
    \begin{minipage}{0.45\textwidth}
        \centering
        \includegraphics[width=\linewidth]{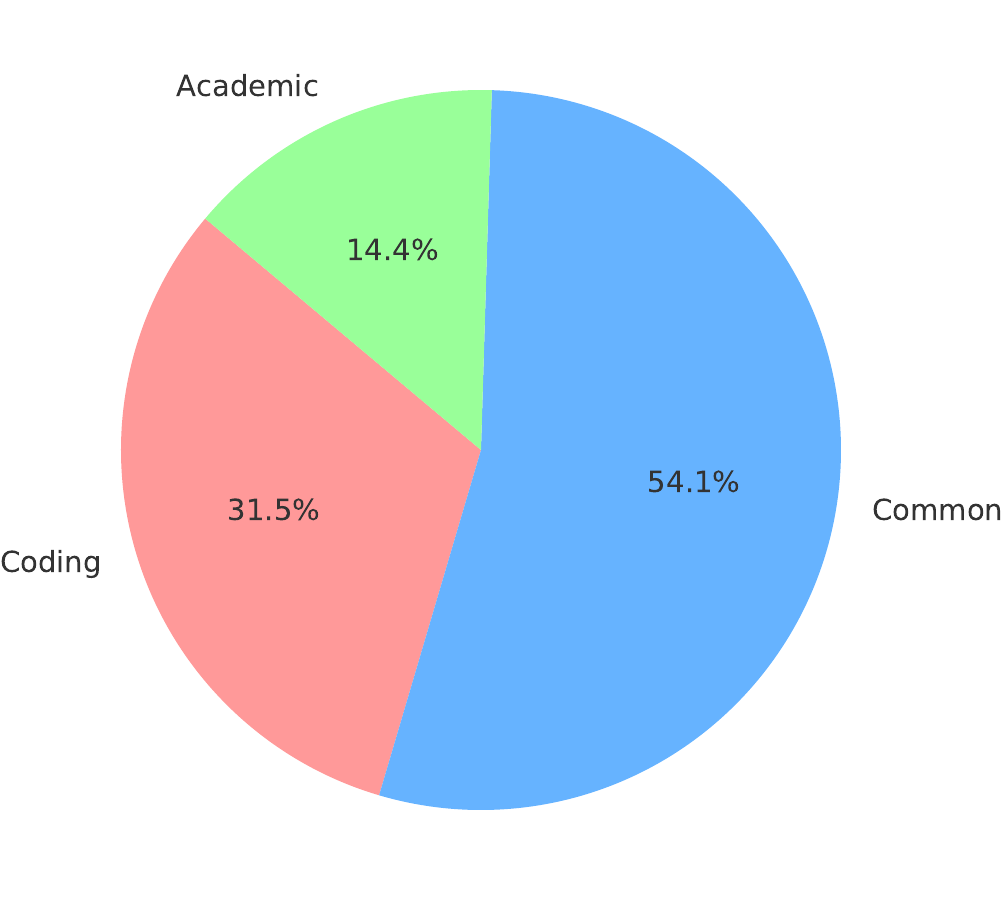} 
        \caption{The proportions of different categories in our whole instruction set (about 5.7 million instructions). Common topics have the highest number of questions, followed by coding questions, with academic questions being the least frequent.}
        \label{fig:pro-inst-set}
    \end{minipage}
\end{figure*}

\section{Instruction Collection}
We aim to study alignment evaluation on generation tasks, such as coding, writing, etc. To better validate our method, we collected a large number of input instructions. We manually filtered datasets from Hugging Face as of June 2023, particularly those in the NLP category. We post-processed the datasets to filter out low-quality instructions as much as possible. We retained all good-quality instructions. We removed instructions that were either too short or too long. We also used the original instructions without tokenization, paraphrasing, etc, to maintain the real distribution of the instructions. After sorting, we keep 37 datasets in total as indicated in Table~\ref{tab:instruction-set} of the Appendix. The training sets of these datasets will be used as instructions. We manually categorized the datasets into three main categories: common, coding, and academic. Common instructions mainly concern everyday matters, such as seeking advice and solving technical problems. All instructions involving coding such as code generation and debugging are classified under the coding category. Lastly, subject-specific instructions, such as science and medicine, are categorized as academic. 

Our cleaned collection consists of around 5.7 million instructions~(see Table~\ref{tab:instruction-set}). We show the diversity of our collection in Fig.~\ref{fig:inst-diversity}, covering different topics. Fig.~\ref{fig:pro-inst-set} shows the proportion of each category. The largest category is common instructions, followed by coding instructions, and then academic instructions. We further calculate the distribution of token counts of the instructions in Fig.~\ref{fig:length-inst}. Most of our instructions are not too long or too short, with the number of tokens mostly between 10 and 20. Some of the instructions are long with a token count larger than 150. We also show some sample instructions in Fig.~\ref{fig:example-instruction}. We keep the original instructions from the source datasets without paraphrasing, to maintain the real distribution of the instructions.

\section{Experiments}
\begin{table}[t]
\caption{Datasets used in our experiments. For data used for instruction tuning, we use GPT-3.5-turbo to generate the reference answers, and GPT-4 is called to provide responses for instructions used in training judge models. 
Dev set is used to search for the optimal $\alpha$ to combine self-evaluation ratings and cosine similarities. For evaluation, we first evaluate methods on our collected instructions, and then AlpacaEval is combined for generalization demonstration. On dev and test sets, we use GPT-4 to generate quality scores for model responses to simulate human ratings.}
\label{tab:data-split}
\centering
\resizebox{0.9\textwidth}{!}{%
\begin{tabular}{lccc}
\textbf{Datasets}        & \textbf{Size} & \textbf{Source of References} & \textbf{Source of Ratings} \\ \midrule
Instruction tuning       & 87k            & ChatGPT     & $-$       \\
Judge model training     & 30k            & GPT-4     & $-$         \\
Dev set                  & 150            & GPT-4     & GPT-4        \\
Eval set 1               & 850            & GPT-4    & GPT-4          \\
Eval set 2 (AlpacaEval) & 805            & GPT-4     & GPT-4       
\end{tabular}%
}
\end{table}

\subsection{Datasets}
We sample a part of the data from our collected instructions to conduct our experiments. 
Table~\ref{tab:data-split} shows the data statistics used in our experiments. Since our collected instructions do not have high-quality answers from the original datasets, and given the prohibitive cost of human labeling, we follow the practice of previous work by using ChatGPT and GPT-4 to simulate human expertise~\cite{vicuna2023,xu2023wizardlm}, which means we use ChatGPT and GPT-4 to generate reference answers for our instructions. We also use GPT-4 to generate quality scores on dev and eval sets to simulate human ratings.

\noindent{\textbf{Instruction Fine-Tuning}} \ From our collected instructions, we sample 87k instructions for training our instruction-following model, where the reference outputs are generated by ChatGPT. Choosing 87k instructions for training follows the setting of Vicuna~\cite{vicuna2023}. 

\noindent{\textbf{Jugde Model Training}} \ We further sample another 30k instructions from our collection for judge model learning, without overlapping with the instruction-tuning set. Different from the instruction-tuning set, for judge model training, the reference answers are from GPT-4, as GPT-4 is substantially better than ChatGPT, which allows us to more accurately validate our algorithm. We set the size of the training set to 30k because calling GPT-4 is too expensive to generate outputs for many more instructions. 30k is a number that we can afford and is a reasonable amount for judge model training (as demonstrated by our experiments). 

We allocate one dataset for instruction tuning and another one for training the judge model, and the two sets do not overlap. However, practically, the two sets can be shared. 
The rationale behind segregating the datasets for instruction tuning and judge model training is primarily due to the high cost associated with calling GPT-4 for generating responses, limiting us to only 30k instructions for judge modeling, but we may need more data to conduct instruction fine-tuning. 
Additionally, for existing instruction-tuned LLMs, we only need the data for judge model training.

\noindent{\textbf{Dev and Eval Sets}} \ We randomly sample 1k for judge model evaluation, where we randomly split the 1k samples into the development set~(150) and test set~(850). Subsequently, for generalization testing, we expand the evaluation by integrating our instructions with AlpacaEval~\cite{alpaca_eval}. 
AlpacaEval is regarded as a cross-domain benchmark that may have a different domain from our collected eval set. 
AlpacaEval includes 805 instructions consisting of diverse tasks, such as coding, writing, reasoning, role-playing, advising, etc. For the dev and eval sets, the reference answers are also from GPT-4. 

We use GPT-4 to generate quality scores for model responses, where we utilize the reference-based estimation since as indicated by \citet{zheng2023judging}, incorporating reference answers can improve the quality of GPT-4's evaluation on reasoning tasks, such as coding problems and mathematical questions. 
As shown in Fig.~\ref{tab:prompt-gpt-eval}, we adopt the template from \citealp{zheng2023judging} and \citealp{DBLP:journals/corr/abs-2305-14314} for GPT-4 evaluation.

\subsection{Setup}

\noindent{\textbf{Instruction Tuning}} \ We use our collected 87k instructions with outputs generated by ChatGPT to fine-tune the Llama-2-13b model~\cite{touvron2023llama}. 
We use LoRA fine-tuning~\cite{hu2021lora}. We set the batch size to 128. We train the model for 3 epochs with a learning rate of 3e-4. For LoRA fine-tuning, we tune the modules of query, key, value, and output projection. The tuned model is named \emph{Ours-13b}.

\noindent{\textbf{Tested Models.}} \ Except for the model Ours-13b, we further study the following open-source models: Vicuna-13b-v1.5~\cite{vicuna2023}, WizardLM-13b-v1.2~\cite{xu2023wizardlm}, Llama-2-13b-Chat, and Llama-2-70b-Chat~\cite{touvron2023llama}. These models are directly used as instruction-tuned models. In total, we have 5 models as the tested models for evaluation. 

\noindent{\textbf{Judge Models.}} \ We train a judge model for each of the tested models. We use 30k instructions for judge model tuning, where the reference answer is from GPT-4. Each tested model samples one response $\bm{y}'$ for each instruction, which creates 30k data points, i.e., ($\bm{x}, \bm{y}',\bm{y}$), for training. Then the model generates the self-rating scores for the sampled responses $\bm{y}'$ by also using the template as in Fig.~\ref{tab:prompt-gpt-eval}. To calculate cosine similarity, we use text-embedding-ada-002 from OpenAI to obtain the embeddings of the answer texts.

We tune Llama-2-13b as the judge models, where LoRA~\cite{hu2021lora} is applied for parameter-efficient tuning. We set the batch size to 128. We train the model for 2 epochs with a learning rate of 3e-4. For LoRA fine-tuning, we tune the modules of query, key, value, and output projection. During self-distillation tuning, the teacher and student losses are jointly optimized on a shared base model. $\beta$ is set to $2$ and $\gamma$ to $0.3$ after parameter selection. The gradient is cut off to back-propagate to the teacher when applying the KL-divergence loss. Experiments were conducted on Nvidia A100 GPUs.

When using the instruction-tuned models to perform self-evaluation, we find that the models may not generate a response that strictly follows the format specified by the prompt in Fig.~\ref{tab:prompt-gpt-eval}. The rating score is required to be in the format of ``\{``rating'': model rating\}''. To deal with this problem, we prompt GPT-3.5-turbo to \emph{extract} the rating score from a response. In practice, we can employ humans or other automatic methods to extract the rating scores.

\begin{table*}[t]
\centering
\large
\caption{
Given GPT-4's status as the leading model, we first rely on GPT-4 to assess the model's response~(using the template in Fig.~\ref{tab:prompt-gpt-eval}), and then calculate the Pearson correlation coefficients (in \%) between various measures with GPT-4's scores. The combined set consists of our eval set and AlpacaEval. Auto-J-13b and UltraRM-13b are supervised models distilled from GPT-4, which are both fine-tuned on Llama-2-13b. PPL and VRO are training-free methods. Each tested model has trained a judge model.
}
\label{tab:main-result}
\resizebox{0.95\textwidth}{!}{%
\begin{tabular}{lcccccc}
 &
  \multicolumn{5}{c}{\textbf{Our 850 test samples}} &
  \multicolumn{1}{l}{} \\
\textbf{Method} &
  \textbf{Ours-13b} &
  \textbf{Vicuna-13b} &
  \textbf{WizardLM-13b} &
  \textbf{Llm2-13b-chat} &
  \textbf{Llm2-70b-chat} &
  \textbf{Avg.} \\ \midrule
\multicolumn{7}{l}{{\ul \textit{\textbf{with reference}}}} \\
Cosine &
  39.75 &
  42.81 &
  40.81 &
  59.04 &
  58.82 &
  48.25 \\
Self-eval &
  44.66 &
  55.13 &
  48.52 &
  40.26 &
  50.70 &
  47.85 \\
Self-eval+cosine &
  53.19 &
  60.77 &
  55.69 &
  64.72 &
  65.51 &
  59.98 \\
GPT-3.5-turbo &
  66.41 &
  66.58 &
  69.96 &
  73.35 &
  75.81 &
  70.42 \\
\ \ \ {$+$ cosine} &
  \textbf{68.33} &
  69.99 &
  \textbf{70.90} &
  \textbf{78.13} &
  \textbf{78.12} &
  73.09 \\
\textsc{Self-J} (ours) &
  66.75 &
  \textbf{70.95} &
  69.56 &
  72.76 &
  71.70 &
  70.34 \\ \midrule
{\ul \textit{\textbf{without reference}}} &
  \multicolumn{1}{l}{\textit{\textbf{}}} &
  \multicolumn{1}{l}{\textit{\textbf{}}} &
  \multicolumn{1}{l}{\textit{\textbf{}}} &
  \multicolumn{1}{l}{\textit{\textbf{}}} &
  \multicolumn{1}{l}{\textit{\textbf{}}} &
  \multicolumn{1}{l}{} \\ 
PPL &
  13.22 &
  13.46 &
  6.47 &
  29.25 &
  -3.99 &
  11.68 \\
VRO &
  45.20 &
  40.03 &
  38.24 &
  40.66 &
  41.47 &
  41.12 \\
Self-eval &
  1.23 &
  15.19 &
  12.75 &
  12.13 &
  15.99 &
  11.46 \\
GPT-3.5-turbo &
  15.21 &
  25.98 &
  19.07 &
  20.05 &
  22.78 &
  20.62 \\
Auto-J-13b &
  37.02 &
  39.68 &
  37.88 &
  53.71 &
  49.43 &
  43.54 \\
UltraRM-13b &
  43.50 &
  44.18 &
  50.68 &
  63.83 &
  62.69 &
  52.98 \\ \midrule
\multicolumn{7}{l}{{\ul \textit{\textbf{Judge models-13b}}}} \\ 
Judge (cosine) &
  39.73 &
  38.78 &
  39.21 &
  61.20 &
  58.06 &
  47.40 \\
Judge (self-eval) &
  45.02 &
  45.14 &
  43.61 &
  48.13 &
  44.57 &
  45.29 \\
\textsc{Self-J} (ours) &
  56.94 &
  56.67 &
  53.10 &
  64.87 &
  61.65 &
  58.65 \\
\ \ \ $-$ \emph{self-distil} &
  50.35 &
  50.75 &
  49.76 &
  62.02 &
  59.91 &
  54.56 \\ \midrule \midrule
 & 
  \multicolumn{6}{c}{\textbf{All 1655 test samples (Our collections + AlpacaEval)}} \\
\textbf{Method} &
  \textbf{Ours-13b} &
  \textbf{Vicuna-13b} &
  \textbf{WizardLM-13b} &
  \textbf{Llm2-13b-chat} &
  \textbf{Llm2-70b-chat} &
  \textbf{Avg.} \\ \midrule
\multicolumn{7}{l}{{\ul \textit{\textbf{with reference}}}} \\
Cosine &
  35.76 &
  41.43 &
  41.41 &
  54.05 &
  54.11 &
  45.35 \\
Self-eval &
  37.87 &
  52.21 &
  42.13 &
  39.96 &
  46.35 &
  43.70 \\
Self-eval+cosine &
  46.68 &
  56.44 &
  48.28 &
  57.65 &
  57.41 &
  53.29 \\
GPT-3.5-turbo &
  63.21 &
  65.63 &
  \textbf{65.56} &
  70.69 &
  \textbf{71.90} &
  67.40 \\
 \ \ \ {$+$ cosine} &
  \textbf{63.70} &
  66.53 &
  63.48 &
  \textbf{73.46} &
  70.89 &
  67.61 \\
\textsc{Self-J} (ours) &
  58.84 &
  \textbf{66.54} &
  65.16 &
  69.39 &
  67.17 &
  65.42 \\ \midrule
{\ul \textit{\textbf{without reference}}} &
  \multicolumn{1}{l}{\textit{\textbf{}}} &
  \multicolumn{1}{l}{\textit{\textbf{}}} &
  \multicolumn{1}{l}{\textit{\textbf{}}} &
  \multicolumn{1}{l}{\textit{\textbf{}}} &
  \multicolumn{1}{l}{\textit{\textbf{}}} &
  \multicolumn{1}{l}{} \\
PPL &
  2.31 &
  -3.30 &
  -2.22 &
  6.41 &
  -2.78 &
  0.08 \\
VRO &
  39.49 &
  36.58 &
  35.57 &
  45.24 &
  44.05 &
  40.19 \\
Self-eval &
  1.41 &
  11.94 &
  14.98 &
  15.05 &
  11.62 &
  11.00 \\
GPT-3.5-turbo &
  18.03 &
  17.21 &
  17.32 &
  18.79 &
  19.11 &
  18.09 \\
Auto-J-13b &
  43.96 &
  40.20 &
  39.36 &
  52.49 &
  48.77 &
  44.96 \\
UltraRM-13b &
  44.87 &
  44.50 &
  49.80 &
  64.54 &
  62.72 &
  53.29 \\ \midrule
\multicolumn{7}{l}{{\ul \textit{\textbf{Judge models-13b}}}} \\
Judge (cosine) &
  31.10 &
  29.86 &
  36.83 &
  58.85 &
  52.14 &
  41.76 \\
Judge (self-eval) &
  44.02 &
  42.39 &
  42.44 &
  53.26 &
  41.92 &
  44.81 \\
\textsc{Self-J} (ours) &
  48.95 &
  51.10 &
  50.72 &
  64.54 &
  59.78 &
  55.02 \\
\ \  \ $-$ \emph{self-distil} &
  41.07 &
  43.16 &
  45.08 &
  60.59 &
  55.06 &
  48.99 \\ 
\end{tabular}%
}
\end{table*}

\subsection{Baselines}
In our evaluation, we consider reference-based and reference-free evaluation. In \emph{reference-based} evaluation, we compare the following baselines:
\begin{itemize}
    \item \textbf{Cosine} \ We calculate the cosine similarity between the embeddings of the model's response and the reference answer, utilizing OpenAI's text-embedding-ada-002 to encode the texts.

    \item \textbf{Self-eval} \  We initiate self-evaluation by the instruction-tuned model itself using the prompt depicted in Fig.~\ref{tab:prompt-gpt-eval}.

    \item \textbf{Self-eval+cosine} \ We merge the scores from self-eval and cosine similarity as outlined in Eq.~\ref{eq:score_combine}, utilizing a development set to determine the optimal hyper-parameter \(\alpha^*\) for this combination.

    \item \textbf{GPT-3.5-turbo} \ Similar to the self-eval method, we employ the prompt detailed in Fig.~\ref{tab:prompt-gpt-eval} to instruct GPT-3.5-turbo to generate quality scores. Additionally, we enhance the scoring by integrating GPT-3.5-turbo's scores with cosine similarities (denoted as ``+ cosine''), applying the optimal value of \(\alpha^*\) identified on the dev set.
\end{itemize}

Then we further experiment with \emph{reference-free} evaluation by comparing to the baselines as follows:
\begin{itemize}
    \item \textbf{PPL} \ We compute the average negative log-likelihood over output tokens. 

    \item \textbf{VRO} \ For the model's output, we generate three additional outputs and compute the sampling variance between the original output and these extra outputs, as elaborated in Eq.~\ref{eq:vro}. We employ OpenAI's text-embedding-ada-002 for text encoding.

    \item \textbf{Auto-J-13b} \ This is an open-source model designed to predict numerical quality scores for alignment evaluation, but unlike ours, it is distilled from GPT-4's evaluation scores~\cite{li2023generative}. This model is also fine-tuned from Llama-2-13b with full parameter fine-tuning. 

    \item \textbf{UltraRM-13b} \ This is a reward model tuned with preference feedback data generated by GPT-4. The model is trained to assign a higher reward to the preferred answer but a lower reward to the rejected answer. The model is also fine-tuned from Llama-2-13b with full-parameter fine-tuning. 
\end{itemize}

Lastly, we report the results of trained judge models. By ablating \textbf{\textsc{Self-J}}, we further train Judge~(cosine) and Judge~(self-eval). Similar to \textsc{Self-J}, both ablated baselines first generate quality scores and then train the judge models with the generated scores. 
\begin{itemize}
    \item \textbf{Judge~(cosine)} relies on the cosine response-reference similarity scores.
    \item \textbf{Jugde~(self-eval)} uses self-evaluation that generates quality scores with reference.
\end{itemize}
For the two ablated baselines, we do not use self-distillation during judge model training. We directly use the generated scores for training.
For \textsc{Self-J}, we report the results of reference-based and reference-free evaluation.

\subsection{Main Results}
We present our main results, the Pearson correlation coefficients with GPT-4, in Table~\ref{tab:main-result}. By analyzing these results, we made the following observations:

\noindent{\textbf{PPL is ineffective for uncertainty estimation in instruction tuning.}} \ Initially, we explore baselines for uncertainty estimation that do not require training. From the results, we can see that PPL is especially poor, exhibiting a low or even negative correlation with GPT-4's evaluation.

\noindent{\textbf{VRO is much better than PPL.}} \ We find that sampling variance reliably estimates alignment, correlating with GPT-4's evaluation. However, it is less effective than tuning-based methods such as Auto-J-13b, UltraRM-13b, and our tuned judge models. Sampling variance is a useful alternative when training a judge model is impractical.

\noindent{\textbf{Cosine similarity and self-evaluation perform well with references.}} \ The methods of \emph{cosine} and \emph{self-eval} both have merits. In reference-based evaluation, we can see that \emph{self-eval} on Vicuna and WizarLM excels, but not on our tuned model or Llama-2-Chat. However, without a reference, \emph{self-eval} becomes much worse than VRO since the effectiveness of self-evaluation is closely related to the model's capabilities. 

\noindent{\textbf{Integrating both methods improves outcomes.}} \ Our method \emph{Self-eval+cosine} outperforms both \emph{cosine} and \emph{self-eval}. It raises correlation by up to 7 points in models like Vicuna and WizardLM. This validates the effectiveness of integrating cosine similarity and self-evaluation. 

\noindent{\textbf{Training judge models proves to be an effective method.}} \ Based on the performance of various judge models, such as Auto-J-13b, UltraRM-13b, and three versions employing different quality scores during training, it is evident that the trained judge models surpass the training-free baseline, \emph{VRO}, in performance. 

\noindent{\textbf{\textsc{Self-J} stands out as the top-performing judge model.}} \ On reference-free evaluation, our approach \textsc{Self-J} outperforms other trained judge models, including those models distilled from GPT-4. It is also much better than GPT-3.5-turbo, where even GPT-3.5-turbo cannot perform well without the reference answers. On reference-based evaluation, we find \textsc{Self-J} to be competitive with GPT-3.5-turbo. \textsc{Self-J} proves to be superior to the method of \emph{Self-eval+cosine}, despite its training data being generated by \emph{Self-eval+cosine}, which is likely attributable to the strong generalization capability of large language models even with noise in the training data.

\begin{figure}
    \centering
    \includegraphics[width=0.75\linewidth]{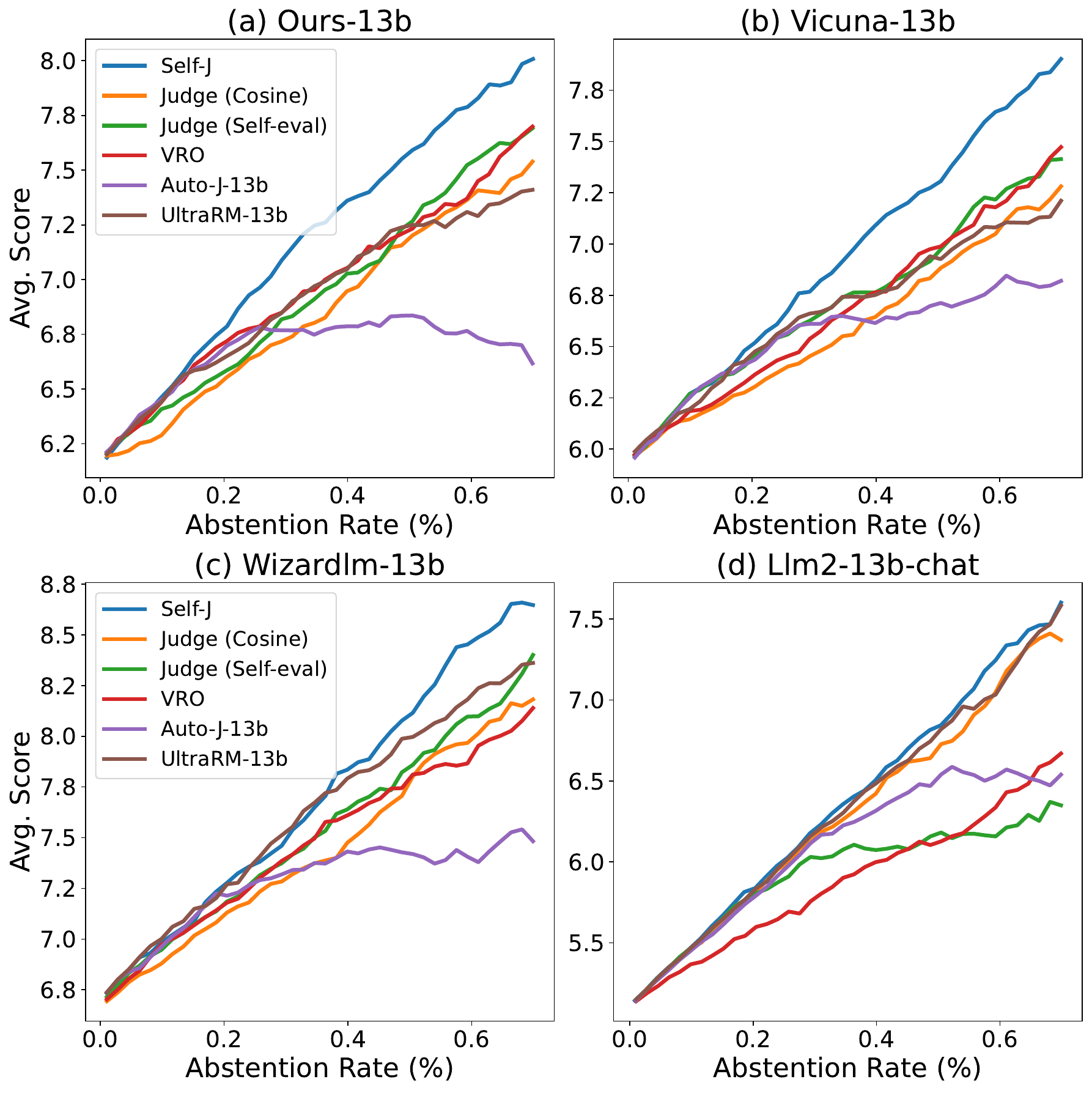}
    \caption{Results for selective instruction following on our collected eval set: average GPT-4 evaluation score versus abstention rate. If the judge model rates a model response below a certain threshold, that response is discarded. By adjusting this threshold, we can generate various combinations of abstention rate and average GPT-4 evaluation score for the model responses that are kept.}
    \label{fig:selective-gen}
\vspace{8mm}
    \centering
    \includegraphics[width=0.7\linewidth]{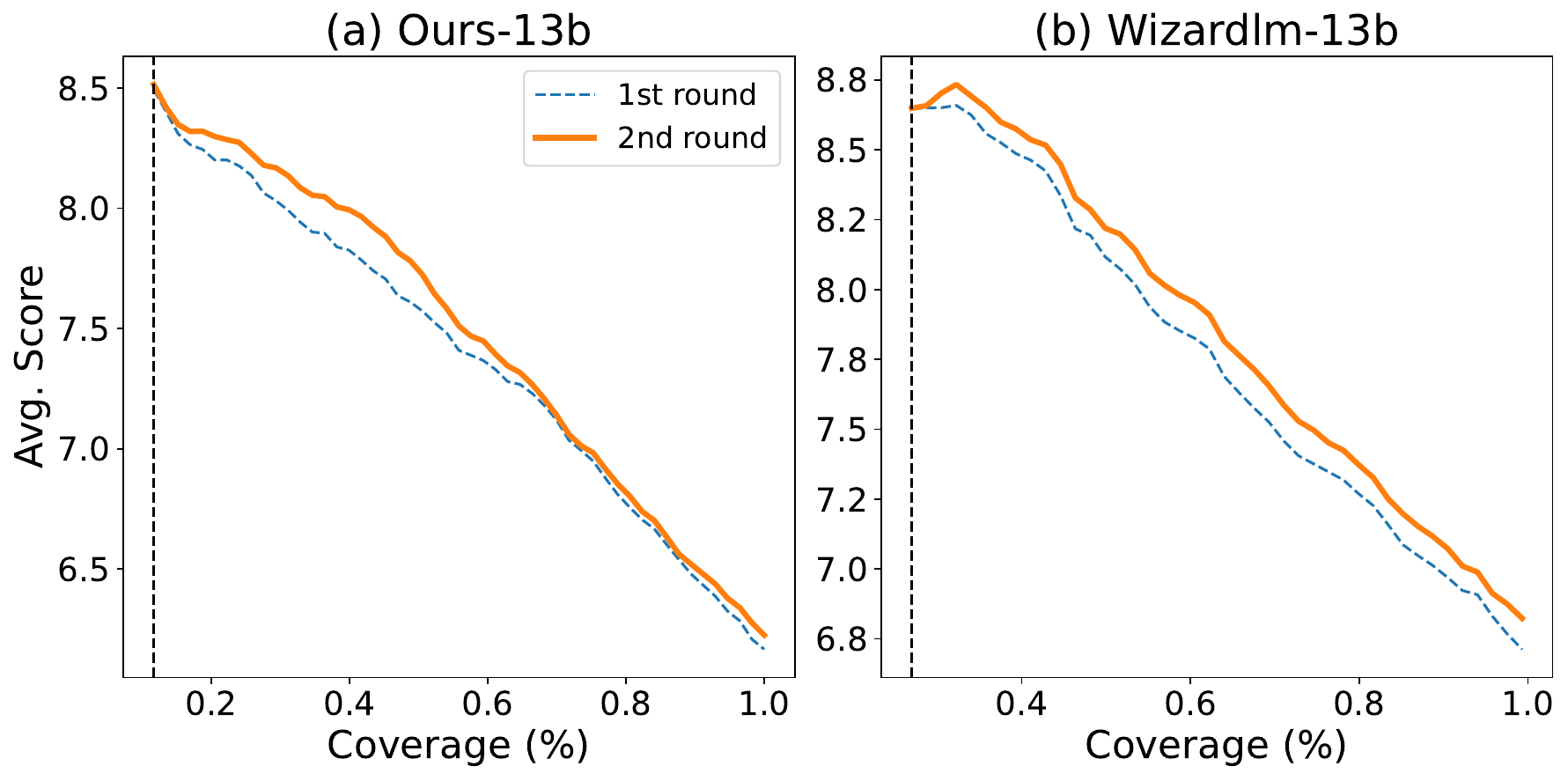}
    \caption{Results for selective refinement on our collected eval set: average GPT-4 evaluation score versus coverage. If a response is rated below a certain threshold by the judge model, it undergoes self-refinement. The refinement process has two stages: first, the model creates feedback $\bm{f}$: $\bm{x} + \bm{y}'_1 + z \rightarrow \bm{f}$. Then, it uses this feedback to refine the initial response $\bm{y}'_1$ into a new response $\bm{y}'_2$, following $\bm{x} + \bm{y}'_1 + \bm{f} \rightarrow \bm{y}'_2$. We plot and compare two curves, one for the first-round response, and one for the second-round response with self-refinement.}
    \label{fig:selective-ref}
\end{figure}

\begin{table}[t]
\centering
\Large
\caption{
Refinement results on our collected test set with GPT-4 evaluation scores (using the template in Fig.~\ref{tab:prompt-gpt-eval} for evaluation). A model refines initial responses using generated feedback for improved second-round outputs. Refinement improves the quality of model responses in each category.  
}
\label{tab:refine}
\resizebox{0.45\textwidth}{!}{%
\begin{tabular}{lcccc} 
         & \multicolumn{2}{c}{\textbf{Ours-13b}} & \multicolumn{2}{c}{\textbf{WizardLM-13b}} \\
         & 1st           & 2nd                   & 1st             & 2nd                     \\ \midrule
Common   & 6.42          & \textbf{6.45}                  & 7.08            & \textbf{7.15}                    \\
Coding   & 5.32          & \textbf{5.33}                  & 5.67            & \textbf{5.84}                    \\
Academic & 7.45          & \textbf{7.53}                  & 7.91            & \textbf{8.01}                    \\ \midrule
All      & 6.17          & \textbf{6.24}         & 6.69            & \textbf{6.85}  \\ 
\end{tabular}%
}
\end{table}

\subsection{Selective Instruction Following and Refinement}
We report more results for selective instruction following. We plot the curve of average GPT-4 evaluation score versus abstention rate, where a model response is discarded if its score rated by the judge model is lower than a threshold, and we vary the threshold to obtain different pairs of abstention rate and average GPT-4 evaluation score of maintained model responses. We show the results of selective instruction following in Fig.~\ref{fig:selective-gen}. For selective instruction following, as expected, the overall performance improves with the exclusion of more responses with lower rating scores from the generation process. Our method \textsc{Self-J} outperforms other baselines by achieving consistently higher GPT-4 scores at different abstention rates. Two ablated models which are Judge (cosine) and Judge (self-eval) perform worse than \textsc{Self-J}. 
We observe that VRO is a strong baseline that can compete with training methods such as UltraRM-13b, Judge (cosine), and Judge (self-eval). We find that Auto-J-13b does not perform well when the abstention rates are high. 

We further study selective refinement, where a model response with a judge model's rating score lower than the threshold will be refined by the model itself. For refinement, the model follows $\bm{x} + \bm{y}'_1 + z \rightarrow \bm{f}$ and $\bm{x} + \bm{y}'_1 + \bm{f} \rightarrow \bm{y}'_2$, where the model needs to generate feedback $\bm{f}$ first, then incorporate the feedback to refine the first-round response $\bm{y}'_1$ to get the second-round response $\bm{y}'_2$. To evaluate the effectiveness of refinement, we plot two curves of average GPT-4 score versus coverage, one for the first-round response and one for the second-round response refined by the model itself. From the results shown in Fig.~\ref{fig:selective-ref}, we can see an improvement by selective refinement. We also present the results of refinement on all responses in Table~\ref{tab:refine} and the refinement process enhances the quality of the model's responses on each category of instructions. Overall, on our 13b model, the average score is improved from 6.17 to 6.24, and on WizardLM-13b, the score is improved from 6.69 to 6.85. Hence, selective refinement enhances model performance.

\begin{figure}[t]
\centering
\includegraphics[width=0.6\columnwidth]{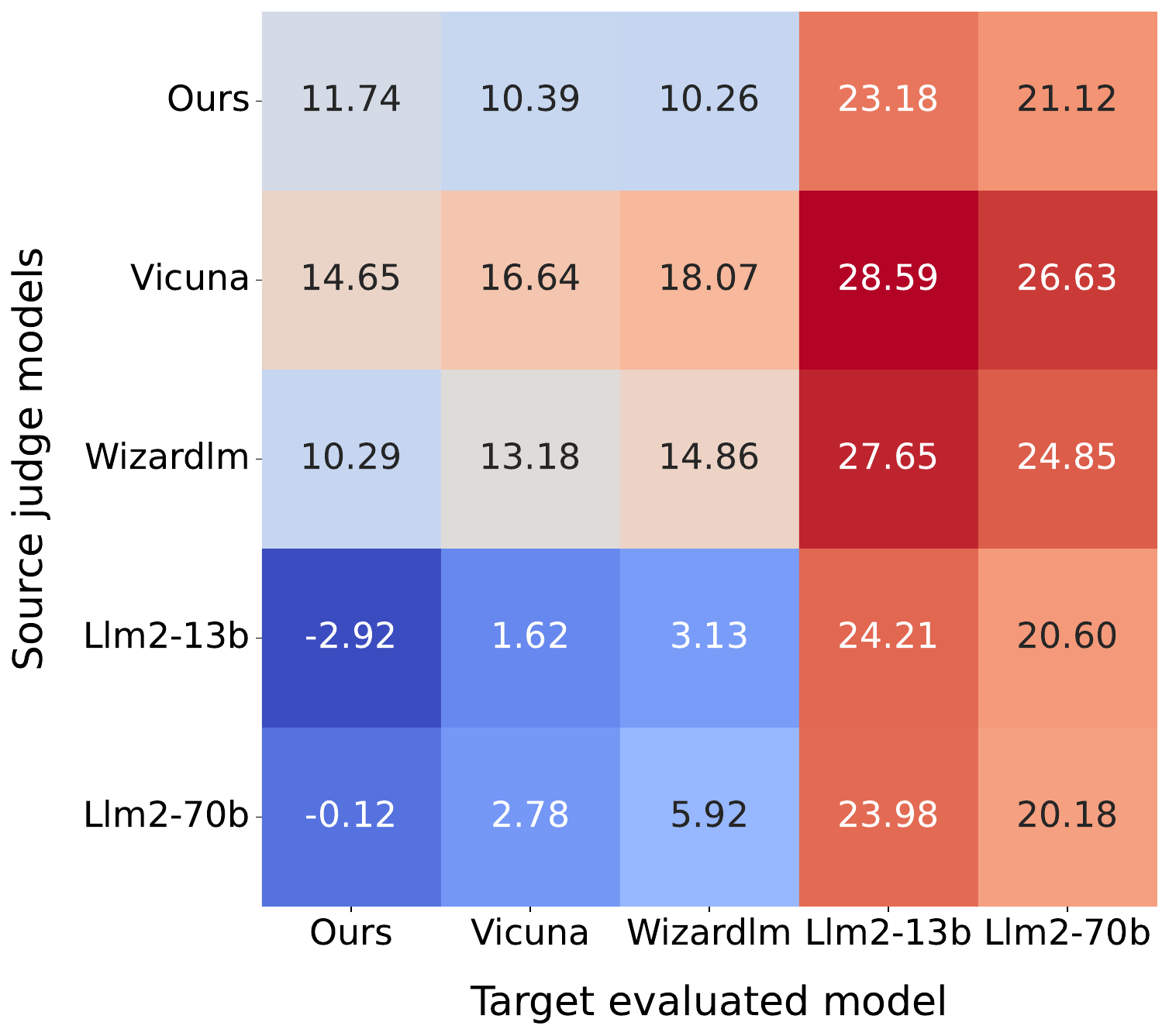} 
\caption{Generalization test on our eval set: judge models from one source assess different target models; \textsc{Self-J}'s Pearson correlation coefficient with GPT-4 minus that of the VRO baseline. VRO is the unsupervised baseline that is free of model training and shows a strong performance. The positive values, i.e., better than VRO, indicate the good generalization ability of judge models. 
}
\label{fig:generalization-judge}
\end{figure}

\begin{table}[t]
\centering
\Large
\caption{System-level Kendall's $\tau$ correlation (in \%) with GPT-4 on AlpacaEval~(version 1), assessed using judge models (\emph{w/o ref.}) across 95 models. For each tested model from the leaderboard of AlpacaEval, we use the judge model (which has been trained for Ours-13b, Vicuna-13b, Wizardlm-13b, Llm2-13b-chat, or Llm2-70b-chat) to rate each response and calculate the average performance on the test set. Using 95 models, we assess the system-level ranking by comparing the ranking derived from the scores of the judge models with that of GPT-4, thereby measuring the correlation between the judge model and GPT-4. See also Fig.~\ref{fig:line_cor_gpt4_vs_judge}.}
\label{tab:system-level-cor}
\resizebox{0.65\columnwidth}{!}{%
\begin{tabular}{lccc} 
Kendall (\%)       & \textbf{\textsc{Self-J}}         & \textbf{Judge (cosine)} & \textbf{Judge (self-eval)} \\ \midrule
Ours-13b      & 50.77          & 41.05          & \textbf{68.51}    \\
Vicuna-13b    & 69.09          & 48.04          & \textbf{70.57}    \\
Wizardlm-13b  & \textbf{68.42} & 38.37          & 59.87             \\
Llm2-13b-chat & \textbf{66.49} & 52.47          & 59.37             \\
Llm2-70b-chat & 62.87          & 52.92          & \textbf{64.21}    \\ \midrule
Avg.          & 63.53          & 46.57          & 64.51         \\ 
\end{tabular}%
}
\end{table}

\begin{table}[t]
\centering
\caption{Best-of-32 sampling results with WizardLM on AlpacaEval using judge models as the reward model. Here, we use the judge models trained for Vicuna-13b since it shows the best performance in Fig.~\ref{fig:generalization-judge} and Table~\ref{tab:system-level-cor}. Our instruction-following model tuned with our collected instructions is comparable to Llama-2-13b-Chat. WizardLM-13b + best-of-32~\big (\textsc{Self-J} and Judge~(self-eval)\big ) outperforms GPT-4 0613 on V2 evaluation.}
\label{tab:best-of-N}
\resizebox{0.7\columnwidth}{!}{%
\begin{tabular}{lcc} 
\textbf{AlpacaEval}                            & \textbf{V1}       & \textbf{V2}       \\ 
Models                                  & (\%) Win & (\%) Win \\ \midrule
gpt4\_turbo                             & 97.70    & 50.00    \\
Yi-34B-Chat                             & 94.08    & 29.66    \\
GPT-4 0613                              & 93.78    & 15.76    \\
GPT 3.5 Turbo 0613                      & 93.42    & 14.13    \\
Claude 2  & 91.36 & 17.19 \\
Claude  & 88.39 & 16.99 \\
LLaMA2 Chat 70B                         & 92.66    & 13.87    \\
UltraLM 13B V2.0 (best-of-16)           & 92.30    & 13.85    \\
PairRM 0.4B+Tulu 2+DPO 13B (best-of-16) & 91.06    & 13.83    \\
Tulu 2+DPO 13B                          & 88.12    & 10.12    \\
Vicuna 13B v1.3                         & 82.11    & 7.14     \\
Vicuna 13B v1.5                         &     -    & 6.70     \\ 
LLaMA2 Chat 13B                         & 81.09    & 7.70     \\
\textbf{Ours-13b}                       & 79.13    & 7.33     \\ \midrule
\textbf{WizardLM-13B-V1.2}              & 89.17    & 12.03    \\
\ \ w/ best-of-32 \textsc{Self-J}       & 92.48    & 15.90     \\
\ \ w/ best-of-32 Judge (cosine)        & 90.87    & 14.47    \\
\ \ w/ best-of-32 Judge (self-eval)     & 93.11    & 17.18   \\ 
\end{tabular}%
}
\end{table}

\begin{figure}[t]
    \centering
    \begin{minipage}{0.62\textwidth}
        \centering
        \includegraphics[width=\linewidth]{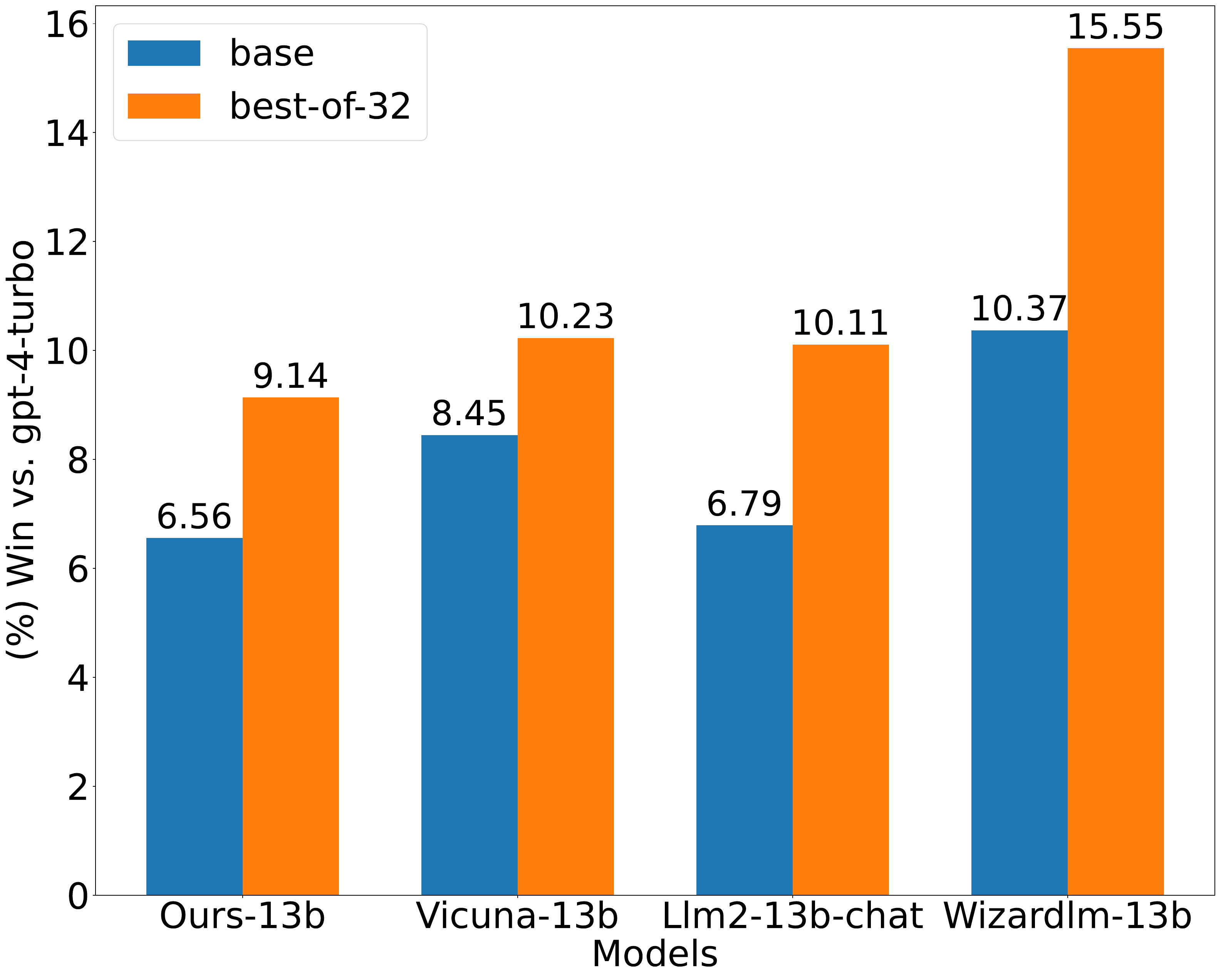} 
        \caption{Results of best-of-32 sampling on each tested model. For every model tested, we use its own judge model~(\textsc{Self-J}) as the reward model. On AlpacaEval, we randomly sample 200 instructions for evaluation. We use v2 in our evaluation. We show the win rate vs. GPT-4-turbo of the model and best-of-32 sampling.
        }
\label{fig:best-of-32-alpaca-eval-part-200}
    \end{minipage}
    \hspace{4pt}
    \begin{minipage}{0.35\textwidth}
        \centering
        \includegraphics[width=\linewidth]{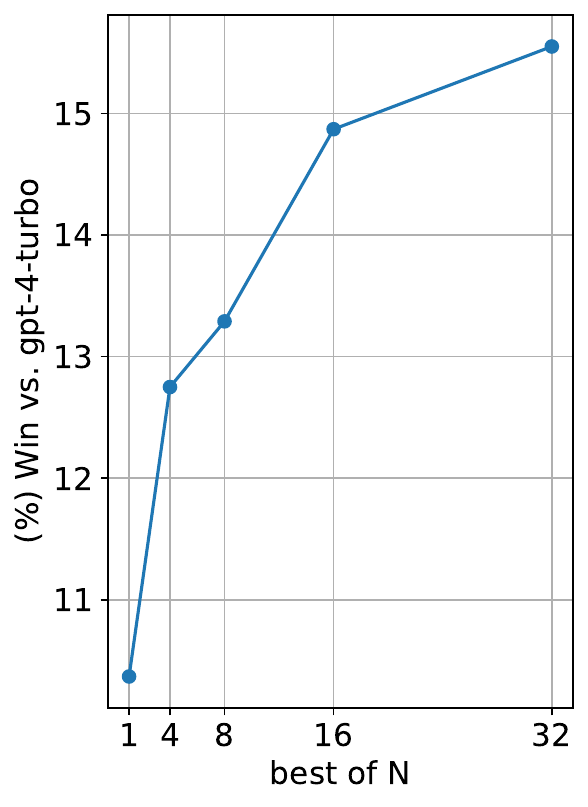} 
        \caption{Results of varying the number of samples based on WizardLM-13b. The setup is the same as Fig.~\ref{fig:best-of-32-alpaca-eval-part-200}. More samples can achieve better performance.}
    \label{fig:best-of-N-vary-N}
    \end{minipage}
\end{figure}

\subsection{Results of Domain Transfer}
\noindent{\textbf{Domain Transfer.}} \ In our study, each judge model is associated with a specific instruction-following model. We examine the judge model's capacity to generalize across various models, which is testing a judge model trained for one source model to evaluate other target models. The results of this generalization test are shown in Fig.~\ref{fig:generalization-judge}. Compared to the VRO baseline, which is a strong baseline without requiring model training, our trained judge models generally exhibit notable improved performance. This highlights the strong generalization ability of the trained judge models. We also find that the judge model trained for Vicuna exhibits the best performance across target models. 

\subsection{Results on AlpacaEval}
The AlpacaEval benchmark has two versions. V1 compares the model with text-davinci-003, but V2 uses the baseline of GPT-4-turbo. 
Here we apply the trained judge models on AlpacaEval by ranking tested models of the leaderboard using the judge models and enhancing the models with best-of-$N$ sampling. 

\noindent{\textbf{Jugde models correlate well with GPT-4 in system-level ranking.}} \ On AlpacaEval v1, we calculate the system-level correlation between judge models and GPT-4. On the 95 evaluated models from the leaderboard, we use judge models to rate the model responses and obtain the average score on the test set. We use the scores measured by judge models to obtain the system-level ranking of the 95 models and then measure the correlation with GPT-4's ranking. 
As shown in Table~\ref{tab:system-level-cor}, of the average result of the five judge models, both \textsc{Self-J} and Judge~(self-eval) show very high correlation with GPT-4. However, Judge~(cosine) is significantly worse than them. Cosine similarity cannot truly understand semantic differences; it is merely a simple measure of similarity between embedded vectors. The judge models trained for different instruction-tuned models do not significantly differ with \textsc{Self-J} or Judge (self-eval). In Table~\ref{tab:ranking_gpt4_judge}, we present all results including the scores measured by the judge model and GPT-4 and the corresponding ranking. Fig.~\ref{fig:line_cor_gpt4_vs_judge} plots the linear fit for win rates by GPT-4 versus \textsc{Self-J} scores. 

\noindent{\textbf{Judge models serve as good reward models.}} \ Additionally, we further use judge models as reward models to enhance the performance of WizardLM-13b with best-of-$32$ sampling. On the test set of AlpacaEval, for each test prompt, we sample 32 responses with WizardLM-13b, each scored by the judge model, with the highest-scoring response selected. Results in Table~\ref{tab:best-of-N} indicate that all versions of the judge model improve performance. We find that the judge model~(self-eval) surpasses \textsc{Self-J}. To explain this, in best-of-$N$ sampling, there is no need for comparing rewards across questions. We only need to compare the rewards for the same question to find the best response. Training a judge model from self-evaluation scores only may be good enough. 
The results of Judge~(self-eval) align with concurrent work, demonstrating that models can self-reward for self-alignment~\cite{yuan2024self}. Our training methodology offers a novel perspective on training reward models without human annotation and utilizing AI models to provide feedback~\cite{lee2023rlaif}. 

We expand best-of-$N$ sampling to more models. For each model, we use its own judge model~(i.e., \textsc{Self-J}) to conduct best-of-$N$ sampling. We select 200 samples from AlpacaEval for evaluation. As indicated by the results of Fig.~\ref{fig:best-of-32-alpaca-eval-part-200}, we find that our judge model can consistently enhance model performance for each tested model. We also study the effects of the number of samples on model improvement. We test the number of samples in \{ 1, 4, 8, 16, 32 \}. The results are shown in Fig.~\ref{fig:best-of-N-vary-N}. We find an increase in performance with a larger number of samples, which further demonstrates the effectiveness of our judge model.

\begin{figure}[t]
\centering
\includegraphics[width=0.8\columnwidth]{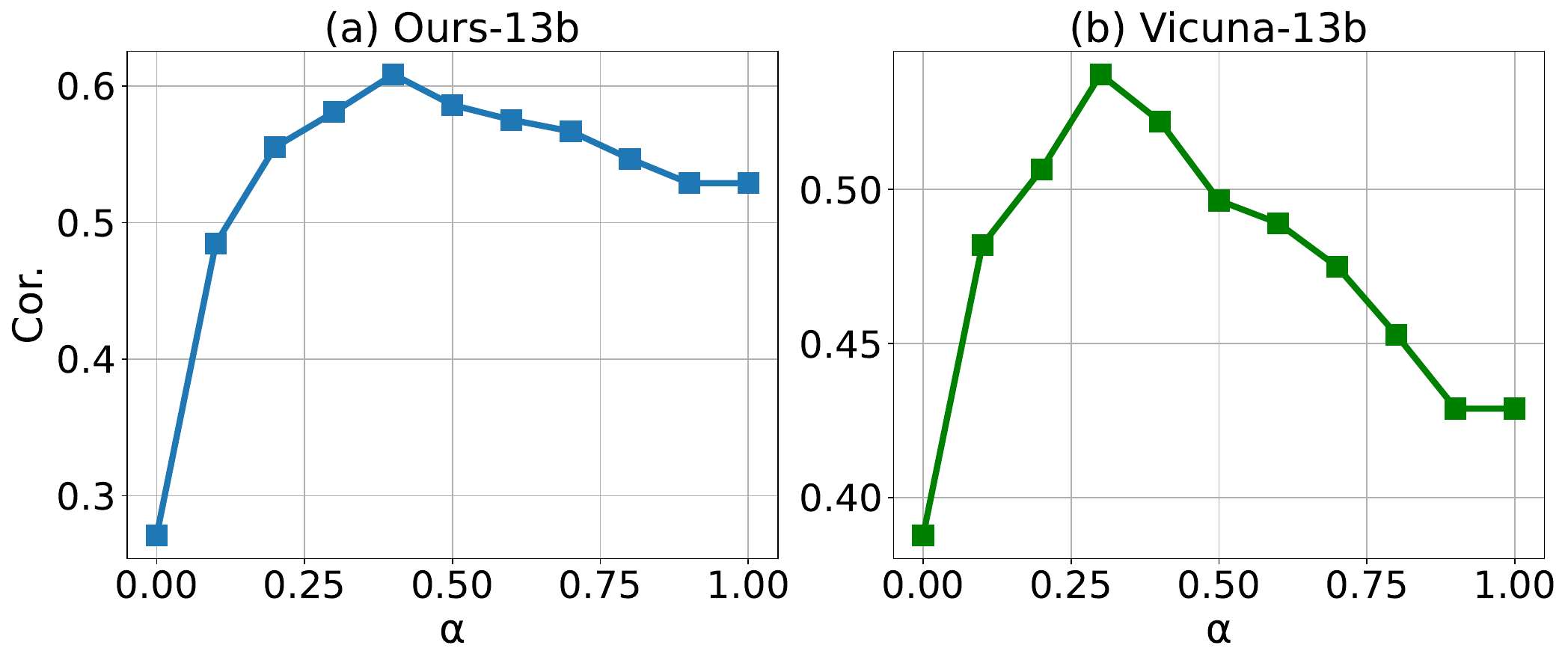} 
\caption{Effects of different values of $\alpha$ (combining self-evaluation and cosine similarity) on Pearson correlation coefficients on the dev set. The two endpoints ($\alpha = 0$ or $1$) of the curve represent the scenarios where only cosine similarity or self-evaluation, respectively, is considered. We find that the curve initially increases, reaches a maximum, and then decreases. Moreover, most $\alpha$ values within the 0--1 range perform better than when $\alpha$ is set to either 0 or 1.}
\label{fig:alpha}
\end{figure}

\subsection{Ablation Study}
\noindent{\textbf{Effect of self-distillation.}} \ In our ablation study, we train a judge model without self-distillation. We directly use generated quality scores to train a judge model using the training loss in Equation \ref{eq:nll}. 
As indicated in Table~\ref{tab:main-result}, the direct training approach ($-$ \emph{self-distil}) results in performance that falls substantially short of the \textsc{Self-J} method under reference-free evaluation. On average, there is a drop of 4$-$6 points in the correlation. 

\noindent{\textbf{Effect of Recalibration for Quality Scores.}} \ Our two ablated baselines Judge (cosine) and Judge (self-eval) use quality scores generated by cosine similarities and self-evaluation respectively. Since neither of them uses self-distillation, we use \textsc{Self-J} without self-distillation for comparison. From the results in Table~\ref{tab:main-result}, our method outperforms the two ablations in terms of correlation scores. This improvement underscores the significance of integrating data sources for quality assessment.

\subsection{Effectiveness of Recalibration for Closed Models}
Here, we study how our method works for the closed model GPT-3.5-turbo. We use GPT-3.5-turbo to generate scores for the five tested models and recalibrate the scores with cosine similarities. We also search for the optimal $\alpha$ to combine the two scores on the development set. As shown in Table~\ref{tab:main-result}, incorporating cosine similarities can further enhance performance. On our collected eval set, our method improves correlation by more than 2 points. This finding demonstrates that our method can also be applied to closed models.

\subsection{Effects of \texorpdfstring{$\alpha$}{alpha}}
We use a held-out development set to search for the optimal $\alpha$ for combining scores of model self-evaluation and cosine similarity. We study how the hyper-parameter $\alpha$ affects the effectiveness of the combination. 
As indicated by the results in Fig.~\ref{fig:alpha}, we find that there exists a considerable range that can reasonably combine the scores of both to demonstrate good composite effects, better than only having self-rating scores or cosine similarities. This finding suggests that our method is robust to the choice of $\alpha$, which is useful in scenarios when obtaining the development set is hard.

\section{Conclusion}

In this paper, we have introduced \textsc{Self-J}, a novel self-training framework designed to enhance the alignment of large language models (LLMs) with human instructions through the development of judge models capable of evaluating the adherence of model outputs to human preferences. 
\textsc{Self-J} adopts a novel self-training method, which operates without the need for human-annotated scores. Our extensive experiments across a variety of open-source models have not only validated the effectiveness of our method but also highlighted its superior performance in comparison to existing baselines, including those distilled from GPT-4, and its competitive edge against GPT-3.5-turbo. Furthermore, the application of \textsc{Self-J} as a reward model has demonstrated significant improvements in model performance, particularly evident in the enhancements achieved with WizardLM-13B-V1.2 under AlpacaEval conditions. These advancements underscore the framework's utility in elevating the quality of model outputs and its contribution to the field of LLMs as a robust tool for alignment evaluation. The high Kendall's tau correlation achieved with GPT-4 in ranking models submitted to AlpacaEval further attests to the reliability and relevance of our judge models in the broader landscape of LLM evaluation. Additionally, our compilation of a collection of large-scale, high-quality instructions for model training and evaluation enriches the resources available for future research, offering a solid foundation for the continued exploration of model alignment and instruction fidelity.

\begin{acknowledgments}
This research is supported by the National Research Foundation, Singapore under its AI Singapore Programme (AISG Award No: AISG2-PhD-2021-08-016[T]). We thank Michael Shieh for his advice on instruction collection, Geonsik Moon for his assistance with instruction collection, and Ruochen Xu for his comments on this work.
\end{acknowledgments}

\starttwocolumn
\bibliography{compling_style}

\onecolumnnew
\clearpage

\appendix

\begin{figure}[t]
\centering
\begin{minipage}{\textwidth}

\begin{tcolorbox}[
  enhanced,
  colback=white,
  colframe=black,
  coltext=black,
  boxsep=0pt,
  arc=0pt,
  outer arc=0pt,
  boxrule=0.8pt,
]

Please act as a precise judge and evaluate the quality of the answer to question. Rate the answer from 0 to 9, where a higher value means a better answer. Please respond with an integer between 0 and 9. 

\vspace{\baselineskip}

[Question]

\{instruction\}
\vspace{\baselineskip}

[Answer]

\{input\}

\end{tcolorbox}
\caption{The template that prompts judge models for reference-free evaluation. The instruction $\bm{x}$ and model response $\bm{y}'$ are combined using the template.}
\label{fig:template-judge-without-refer}
\end{minipage}
\vspace{2mm}

\begin{minipage}{\textwidth}
    \begin{tcolorbox}[
  enhanced,
  colback=white,
  colframe=black,
  coltext=black,
  boxsep=0pt,
  arc=0pt,
  outer arc=0pt,
  boxrule=0.8pt, 
]

Please act as a precise judge and evaluate the quality of the answer to question. Rate the answer from 0 to 9, where a higher value means a better answer. Please refer to the reference answer to make your judgment. Respond with an integer between 0 and 9. 

\vspace{\baselineskip}

[Question]

\{instruction\}
\vspace{\baselineskip}

[Reference]

\{reference\}
\vspace{\baselineskip}

[Answer]

\{input\}

\end{tcolorbox}
\caption{The template that prompts judge models for reference-based evaluation. The instruction $\bm{x}$, reference answer $\bm{y}$, and model response $\bm{y}'$ are combined using the template.}
\label{fig:template-judge-with-refer}
\end{minipage}

\end{figure}

\begin{table}[t]
\centering
\Huge
\caption{The statistics of our collected instructions, including the category, data source, and number of instructions. Our collection has 37 datasets and around 5.7 million instructions.}
\label{tab:instruction-set}
\resizebox{\textwidth}{!}{%
\begin{tabular}{r|l|l}
\multicolumn{1}{l|}{}               & \textbf{Dataset}                                & \textbf{Count} \\ \hline
\multirow{14}{*}{\textbf{Common}}   & embedding-data\_PAQ\_pairs                      & 555168         \\
                                    & embedding-data\_WikiAnswers\_train              & 395392         \\
                                    & BeIR\_cqadupstack-generated-queries             & 376675         \\
                                    & ms\_marco                                       & 346651         \\
                                    & koutch\_yahoo\_answers\_topics                  & 307245         \\
                                    & totuta\_youtube\_subs\_howto100M                & 283431         \\
                                    & quora                                           & 254391         \\
 & flax-sentence-embeddings\_stackexchange\_titlebody\_best\_and\_down\_voted\_answer           & 242382 \\
                                    & LLukas22\_lfqa\_preprocessed                    & 238427         \\
                                    & koutch\_yahoo\_answers\_qa                      & 73304          \\
                                    & AmazonScience\_mintaka                          & 17232          \\
                                    & common\_questions\_piqa                         & 14814          \\
                                    & b-mc2\_wikihow\_lists                           & 6055           \\
                                    & launch\_open\_question\_type                    & 1261           \\ \hline
\multirow{11}{*}{\textbf{Coding}}   & pacovaldez\_stackoverflow                       & 834728         \\
                                    & koutch\_stackoverflow\_python                   & 672414         \\
                                    & koutch\_staqc                                   & 224585         \\
                                    & neulab\_conala                                  & 42970          \\
                                    & sedthh\_ubuntu\_dialogue\_qa                    & 12709          \\
                                    & BeIR\_cqadupstack-generated-queries\_coding     & 10466          \\
                                    & neulab\_tldr                                    & 5985           \\
 & flax-sentence-embeddings\_stackexchange\_titlebody\_best\_and\_down\_voted\_answer\_coding   & 4541   \\
                                    & koutch\_yahoo\_answers\_topics\_coding          & 4063           \\
                                    & quora\_coding                                   & 2051           \\
                                    & mbpp                                            & 690            \\ \hline

\multirow{12}{*}{\textbf{Academic}} & pubmed\_qa                                      & 269833         \\
                                    & medical\_dialog                                 & 225710         \\
                                    & medmcqa                                         & 144248         \\
                                    & flax-sentence-embeddings\_stackexchange\_math   & 107738         \\
                                    & medalpaca\_medical\_meadow\_medical\_flashcards & 29823          \\
                                    & danielpark\_MQuAD-v1                            & 15733          \\
                                    & qasc                                            & 13072          \\
                                    & sciq                                            & 10126          \\
 & flax-sentence-embeddings\_stackexchange\_titlebody\_best\_and\_down\_voted\_answer\_academic & 5408   \\
                                    & cannin\_biostars\_qa                            & 2537           \\
                                    & covid\_qa\_deepset                              & 1451           \\
                                    & medical\_questions\_pairs                       & 1103           \\ \hline
\textbf{All}                        & 37                                              & 5754412        \\ 
\end{tabular}%
}
\end{table}

\begin{figure*}[t]
\setlength{\abovecaptionskip}{0.2cm}
\setlength{\belowcaptionskip}{-0.3cm}
\centering
\includegraphics[width=\textwidth]{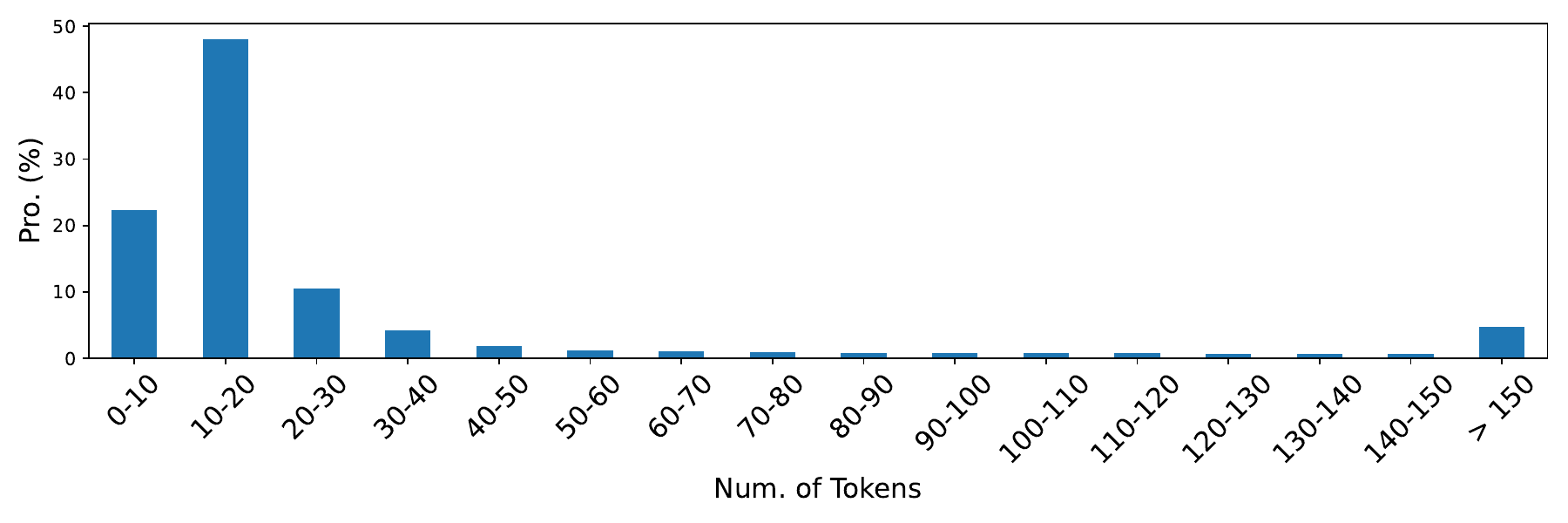} 
\caption{The distribution of token counts of instructions in our collected set, based on around 6 million instructions. Most of the instructions contain 10 to 20 tokens, but a significant number of instructions are longer.}
\label{fig:length-inst}
\end{figure*}

\begin{figure}
\begin{tcolorbox}[
  enhanced,
  colback=white,
  colframe=black,
  coltext=black,
  boxsep=0pt,
  arc=0pt,
  outer arc=0pt,
  boxrule=0.8pt,
]

\begin{enumerate}[itemsep=0pt, topsep=0pt]
    \item \textbf{(Biomedical-PubMed\_abstracts)} \ Is expression of eukaryotic initiation factor 3f associated with prognosis in gastric carcinomas? 

    \item \textbf{(embedding-data\_WikiAnswers\_train)} \ Was the Anaconda plan of the civil war successful?

    \item \textbf{(coding-StackOverflow)} \ Configure sidekiq to work without brocker in development environment 

    \item \textbf{(common-reddit)} \ how are commercial airplanes supplied with electricity? 

    \item \textbf{ (common-Wikipedia)} \ how many stems are formed on each root in akkadian verbs

    \item \textbf{(common-YouTube)} \ How to make the classic sloe gin fizz

    \item \textbf{(coding-StackOverflow-python)} \ publish on facebook page from python cron job 

    \item \textbf{(MedDialog-icliniq\_healthcaremagic\_healthtap)} \ What causes constant vomiting and green mucus in eyes of an infant? 

    \item \textbf{(coding-StackOverflow-python)} \ I'm trying to get the shift-jis character code from a unicode string. I'm not
really that knowledgable in python, but here is what I have tried so far:

    \#!/usr/bin/env python
    
    \# -*- coding: utf-8 -*-
    
    from struct import *
    
    data=``è<87><8d>''
    
    udata=data.decode(``utf-8'')
    
    data=udata.encode(``shift-jis'').decode(``shift-jis'')
    
    code=unpack(data, ``Q'')
    
    print code

But I get an `UnicodeEncodeError: `ascii' codec can't encode character
u`\u81cd' in position 0: ordinal not in range(128)' error. The string is
always a single character. 

\item \textbf{(Math-StackExchange)} \ Please give me some advice on references of complex geometry Recently I am reading complex geometry and preparing for my complex geometry exam. Our lectures book is so disorder and brief that I have to consult Wikipedia and math-overflow. I need some materials of complex geometry such as almost complex manifolds, Kähler manifolds, complex and holomorphic vector bundles, Hodge theory, Chern classes and sheaf theory. So can you recommend some complex geometry lectures or books to me?

Any help will be greatly appreciated!

Thanks in advance! 
    
\end{enumerate}
\end{tcolorbox}
\caption{Sample instructions from our collection.}
\label{fig:example-instruction}
\end{figure}

\begin{figure*}[t]
    \centering
    \includegraphics[width=0.8\linewidth]{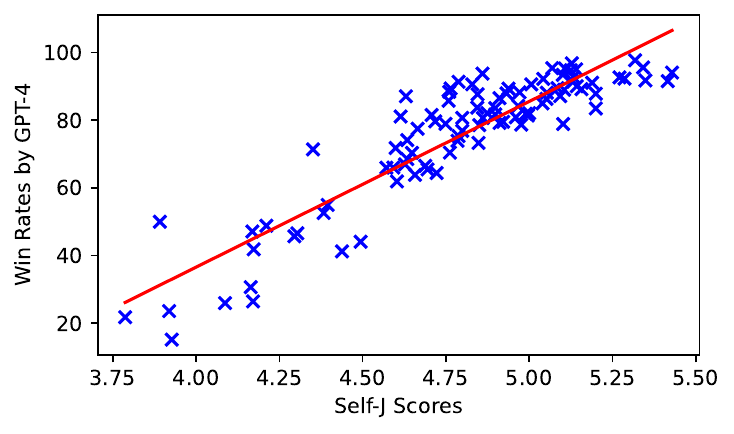}
    \caption{Scatter plot with linear fit for win rates by GPT-4 versus \textsc{Self-J} scores. We use the judge model trained for Vicuna-13b. There are 95 models from AlpacaEval.}
    \label{fig:line_cor_gpt4_vs_judge}
\end{figure*}

\begin{longtable}[b]
{|p{3cm}|>{\centering\arraybackslash}p{2cm}|>{\centering\arraybackslash}p{2cm}|>{\centering\arraybackslash}p{1cm}|>{\centering\arraybackslash}p{1.5cm}|>{\centering\arraybackslash}p{0.5cm}|}
\caption{Full results of system-level scores measured by \textsc{Self-J} and GPT-4. GPT-4 provides the win rates of the tested model against text-davinci003. We also provide the rankings created by the judge model and GPT-4 respectively. $\Delta$ is calculated by subtracting the ranking provided by GPT-4 from the ranking determined by the judge model. The judge model is trained for Vicuna-13b. There are 95 models in total.}
\label{tab:ranking_gpt4_judge}
\\
\hline
                     & \multicolumn{2}{c|}{\textbf{Scores}}              & \multicolumn{2}{c|}{\textbf{Ranking}}                            & \multicolumn{1}{l|}{} \\
\textbf{Model} & \textbf{Self-J} & \textbf{GPT-4} & \textbf{Self-J} & \textbf{GPT-4} & \textbf{$\Delta$} \\ \hline
\endfirsthead

\hline
& \multicolumn{2}{c|}{\textbf{Scores}}              & \multicolumn{2}{c|}{\textbf{Ranking}}                            & \multicolumn{1}{l|}{} \\
\textbf{Model} & \textbf{Self-J} & \textbf{GPT-4} & \textbf{Self-J} & \textbf{GPT-4} & \textbf{$\Delta$} \\ \hline
\endhead

\multicolumn{6}{|r|}{{Continued on next page}} \\ \hline
\endfoot

\hline
\endlastfoot

Yi-34B-Chat                  & \multicolumn{1}{c|}{5.429} & 94.08 & \multicolumn{1}{c|}{1}  & 8  & -7                    \\ \hline
ultralm-13b-best-of-16       & \multicolumn{1}{c|}{5.416} & 91.54 & \multicolumn{1}{c|}{2}  & 16 & -14                   \\ \hline
xwinlm-13b-v0.1              & \multicolumn{1}{c|}{5.349} & 91.76 & \multicolumn{1}{c|}{3}  & 15 & -12                   \\ \hline
xwinlm-70b-v0.1              & \multicolumn{1}{c|}{5.342} & 95.57 & \multicolumn{1}{c|}{4}  & 3  & 1                     \\ \hline
gpt4\_turbo                  & \multicolumn{1}{c|}{5.318} & 97.70 & \multicolumn{1}{c|}{5}  & 1  & 4                     \\ \hline
ultralm-13b-v2.0-best-of-16  & \multicolumn{1}{c|}{5.286} & 92.30 & \multicolumn{1}{c|}{6}  & 13 & -7                    \\ \hline
llama-2-70b-chat-hf          & \multicolumn{1}{c|}{5.271} & 92.66 & \multicolumn{1}{c|}{7}  & 12 & -5                    \\ \hline
recycled-wizardlm-7b-v2.0    & \multicolumn{1}{c|}{5.200} & 83.48 & \multicolumn{1}{c|}{8}  & 43 & -35                   \\ \hline
xwinlm-7b-v0.1               & \multicolumn{1}{c|}{5.200} & 87.83 & \multicolumn{1}{c|}{9}  & 32 & -23                   \\ \hline
pairrm-tulu-2-13b            & \multicolumn{1}{c|}{5.189} & 91.06 & \multicolumn{1}{c|}{10} & 19 & -9                    \\ \hline
wizardlm-13b-v1.2            & \multicolumn{1}{c|}{5.157} & 89.17 & \multicolumn{1}{c|}{11} & 26 & -15                   \\ \hline
deita-7b-v1.0                & \multicolumn{1}{c|}{5.143} & 90.06 & \multicolumn{1}{c|}{12} & 22 & -10                   \\ \hline
tulu-2-dpo-70b               & \multicolumn{1}{c|}{5.141} & 95.03 & \multicolumn{1}{c|}{13} & 6  & 7                     \\ \hline
Mistral-7B-Instruct-v0.2     & \multicolumn{1}{c|}{5.133} & 92.78 & \multicolumn{1}{c|}{14} & 11 & 3                     \\ \hline
cut-13b                      & \multicolumn{1}{c|}{5.132} & 91.36 & \multicolumn{1}{c|}{15} & 17 & -2                    \\ \hline
mistral-medium               & \multicolumn{1}{c|}{5.128} & 96.83 & \multicolumn{1}{c|}{16} & 2  & 14                    \\ \hline
Mixtral-8x7B-Instruct-v0.1   & \multicolumn{1}{c|}{5.125} & 94.78 & \multicolumn{1}{c|}{17} & 7  & 10                    \\ \hline
gpt4                         & \multicolumn{1}{c|}{5.107} & 95.28 & \multicolumn{1}{c|}{18} & 5  & 13                    \\ \hline
vicuna-33b-v1.3              & \multicolumn{1}{c|}{5.105} & 88.99 & \multicolumn{1}{c|}{19} & 27 & -8                    \\ \hline
recycled-wizardlm-7b-v1.0    & \multicolumn{1}{c|}{5.102} & 78.88 & \multicolumn{1}{c|}{20} & 58 & -38                   \\ \hline
pairrm-zephyr-7b-beta        & \multicolumn{1}{c|}{5.101} & 93.41 & \multicolumn{1}{c|}{21} & 10 & 11                    \\ \hline
openchat-v2-w-13b            & \multicolumn{1}{c|}{5.093} & 87.13 & \multicolumn{1}{c|}{22} & 34 & -12                   \\ \hline
openchat-v3.1-13b            & \multicolumn{1}{c|}{5.085} & 89.49 & \multicolumn{1}{c|}{23} & 23 & 0                     \\ \hline
pairrm-tulu-2-70b            & \multicolumn{1}{c|}{5.069} & 95.40 & \multicolumn{1}{c|}{24} & 4  & 20                    \\ \hline
tulu-2-dpo-13b               & \multicolumn{1}{c|}{5.053} & 88.12 & \multicolumn{1}{c|}{25} & 30 & -5                    \\ \hline
wizardlm-13b-v1.1            & \multicolumn{1}{c|}{5.052} & 86.32 & \multicolumn{1}{c|}{26} & 37 & -11                   \\ \hline
LMCocktail-10.7B-v1          & \multicolumn{1}{c|}{5.042} & 92.22 & \multicolumn{1}{c|}{27} & 14 & 13                    \\ \hline
openchat-v2-13b              & \multicolumn{1}{c|}{5.040} & 84.97 & \multicolumn{1}{c|}{28} & 39 & -11                   \\ \hline
zephyr-7b-beta               & \multicolumn{1}{c|}{5.006} & 90.60 & \multicolumn{1}{c|}{29} & 21 & 8                     \\ \hline
llama-2-chat-7b-evol70k-neft & \multicolumn{1}{c|}{5.000} & 82.09 & \multicolumn{1}{c|}{30} & 45 & -15                   \\ \hline
phi-2-dpo                    & \multicolumn{1}{c|}{4.996} & 81.37 & \multicolumn{1}{c|}{31} & 49 & -18                   \\ \hline
platolm-7b                   & \multicolumn{1}{c|}{4.990} & 81.94 & \multicolumn{1}{c|}{32} & 46 & -14                   \\ \hline
opencoderplus-15b            & \multicolumn{1}{c|}{4.976} & 78.70 & \multicolumn{1}{c|}{33} & 59 & -26                   \\ \hline
causallm-14b                 & \multicolumn{1}{c|}{4.971} & 88.26 & \multicolumn{1}{c|}{34} & 29 & 5                     \\ \hline
tulu-2-dpo-7b                & \multicolumn{1}{c|}{4.967} & 84.22 & \multicolumn{1}{c|}{35} & 40 & -5                    \\ \hline
openchat-13b                 & \multicolumn{1}{c|}{4.960} & 80.87 & \multicolumn{1}{c|}{36} & 51 & -15                   \\ \hline
evo-v2-7b                    & \multicolumn{1}{c|}{4.938} & 89.35 & \multicolumn{1}{c|}{37} & 25 & 12                    \\ \hline
humpback-llama2-70b          & \multicolumn{1}{c|}{4.932} & 87.94 & \multicolumn{1}{c|}{38} & 31 & 7                     \\ \hline
openchat8192-13b             & \multicolumn{1}{c|}{4.922} & 79.54 & \multicolumn{1}{c|}{39} & 55 & -16                   \\ \hline
evo-7b                       & \multicolumn{1}{c|}{4.911} & 79.20 & \multicolumn{1}{c|}{40} & 56 & -16                   \\ \hline
openbuddy-llama-65b-v8       & \multicolumn{1}{c|}{4.911} & 86.53 & \multicolumn{1}{c|}{41} & 36 & 5                     \\ \hline
ultralm-13b-v2.0             & \multicolumn{1}{c|}{4.898} & 83.60 & \multicolumn{1}{c|}{42} & 42 & 0                     \\ \hline
gpt35\_turbo\_instruct       & \multicolumn{1}{c|}{4.898} & 81.71 & \multicolumn{1}{c|}{43} & 47 & -4                    \\ \hline
vicuna-13b-v1.3              & \multicolumn{1}{c|}{4.874} & 82.11 & \multicolumn{1}{c|}{44} & 44 & 0                     \\ \hline
ultralm-13b                  & \multicolumn{1}{c|}{4.861} & 80.64 & \multicolumn{1}{c|}{45} & 53 & -8                    \\ \hline
gpt4\_0613                   & \multicolumn{1}{c|}{4.860} & 93.78 & \multicolumn{1}{c|}{46} & 9  & 37                    \\ \hline
minichat-1.5-3b              & \multicolumn{1}{c|}{4.850} & 78.55 & \multicolumn{1}{c|}{47} & 60 & -13                   \\ \hline
airoboros-33b                & \multicolumn{1}{c|}{4.848} & 73.29 & \multicolumn{1}{c|}{48} & 66 & -18                   \\ \hline
openbuddy-llama2-70b-v10.1   & \multicolumn{1}{c|}{4.845} & 87.67 & \multicolumn{1}{c|}{49} & 33 & 16                    \\ \hline
humpback-llama-65b           & \multicolumn{1}{c|}{4.844} & 83.71 & \multicolumn{1}{c|}{50} & 41 & 9                     \\ \hline
cohere                       & \multicolumn{1}{c|}{4.829} & 90.62 & \multicolumn{1}{c|}{51} & 20 & 31                    \\ \hline
vicuna-7b-v1.3               & \multicolumn{1}{c|}{4.799} & 76.84 & \multicolumn{1}{c|}{52} & 62 & -10                   \\ \hline
openbuddy-falcon-40b-v9      & \multicolumn{1}{c|}{4.799} & 80.70 & \multicolumn{1}{c|}{53} & 52 & 1                     \\ \hline
claude-2                     & \multicolumn{1}{c|}{4.788} & 91.36 & \multicolumn{1}{c|}{54} & 18 & 36                    \\ \hline
wizardlm-13b                 & \multicolumn{1}{c|}{4.788} & 75.31 & \multicolumn{1}{c|}{55} & 63 & -8                    \\ \hline
airoboros-65b                & \multicolumn{1}{c|}{4.785} & 73.91 & \multicolumn{1}{c|}{56} & 65 & -9                    \\ \hline
gpt-3.5-turbo-0301           & \multicolumn{1}{c|}{4.764} & 89.37 & \multicolumn{1}{c|}{57} & 24 & 33                    \\ \hline
vicuna-13b                   & \multicolumn{1}{c|}{4.762} & 70.43 & \multicolumn{1}{c|}{58} & 69 & -11                   \\ \hline
claude                       & \multicolumn{1}{c|}{4.760} & 88.39 & \multicolumn{1}{c|}{59} & 28 & 31                    \\ \hline
zephyr-7b-alpha              & \multicolumn{1}{c|}{4.758} & 85.76 & \multicolumn{1}{c|}{60} & 38 & 22                    \\ \hline
claude2-alpaca-13b           & \multicolumn{1}{c|}{4.749} & 78.93 & \multicolumn{1}{c|}{61} & 57 & 4                     \\ \hline
vicuna-7b                    & \multicolumn{1}{c|}{4.722} & 64.41 & \multicolumn{1}{c|}{62} & 77 & -15                   \\ \hline
gemini-pro                   & \multicolumn{1}{c|}{4.718} & 79.66 & \multicolumn{1}{c|}{63} & 54 & 9                     \\ \hline
openbuddy-llama-30b-v7.1     & \multicolumn{1}{c|}{4.707} & 81.55 & \multicolumn{1}{c|}{64} & 48 & 16                    \\ \hline
nous-hermes-13b              & \multicolumn{1}{c|}{4.695} & 65.47 & \multicolumn{1}{c|}{65} & 76 & -11                   \\ \hline
oasst-rlhf-llama-33b         & \multicolumn{1}{c|}{4.688} & 66.52 & \multicolumn{1}{c|}{66} & 73 & -7                    \\ \hline
openbuddy-llama2-13b-v11.1   & \multicolumn{1}{c|}{4.665} & 77.49 & \multicolumn{1}{c|}{67} & 61 & 6                     \\ \hline
baize-v2-7b                  & \multicolumn{1}{c|}{4.657} & 63.85 & \multicolumn{1}{c|}{68} & 78 & -10                   \\ \hline
openbuddy-falcon-7b-v6       & \multicolumn{1}{c|}{4.648} & 70.36 & \multicolumn{1}{c|}{69} & 70 & -1                    \\ \hline
phi-2-sft                    & \multicolumn{1}{c|}{4.634} & 68.53 & \multicolumn{1}{c|}{70} & 71 & -1                    \\ \hline
jina-chat                    & \multicolumn{1}{c|}{4.634} & 74.13 & \multicolumn{1}{c|}{71} & 64 & 7                     \\ \hline
claude-2.1                   & \multicolumn{1}{c|}{4.630} & 87.08 & \multicolumn{1}{c|}{72} & 35 & 37                    \\ \hline
baize-v2-13b                 & \multicolumn{1}{c|}{4.627} & 66.96 & \multicolumn{1}{c|}{73} & 72 & 1                     \\ \hline
llama-2-13b-chat-hf          & \multicolumn{1}{c|}{4.614} & 81.09 & \multicolumn{1}{c|}{74} & 50 & 24                    \\ \hline
alpaca-7b-neft               & \multicolumn{1}{c|}{4.603} & 61.92 & \multicolumn{1}{c|}{75} & 79 & -4                    \\ \hline
guanaco-65b                  & \multicolumn{1}{c|}{4.599} & 71.80 & \multicolumn{1}{c|}{76} & 67 & 9                     \\ \hline
minotaur-13b                 & \multicolumn{1}{c|}{4.592} & 66.02 & \multicolumn{1}{c|}{77} & 74 & 3                     \\ \hline
guanaco-33b                  & \multicolumn{1}{c|}{4.571} & 65.96 & \multicolumn{1}{c|}{78} & 75 & 3                     \\ \hline
alpaca-farm-ppo-sim-gpt4-20k & \multicolumn{1}{c|}{4.494} & 44.10 & \multicolumn{1}{c|}{79} & 87 & -8                    \\ \hline
alpaca-farm-ppo-human        & \multicolumn{1}{c|}{4.438} & 41.24 & \multicolumn{1}{c|}{80} & 89 & -9                    \\ \hline
oasst-sft-llama-33b          & \multicolumn{1}{c|}{4.395} & 54.97 & \multicolumn{1}{c|}{81} & 80 & 1                     \\ \hline
guanaco-13b                  & \multicolumn{1}{c|}{4.383} & 52.61 & \multicolumn{1}{c|}{82} & 81 & 1                     \\ \hline
llama-2-7b-chat-hf           & \multicolumn{1}{c|}{4.351} & 71.37 & \multicolumn{1}{c|}{83} & 68 & 15                    \\ \hline
guanaco-7b                   & \multicolumn{1}{c|}{4.304} & 46.58 & \multicolumn{1}{c|}{84} & 85 & -1                    \\ \hline
falcon-40b-instruct          & \multicolumn{1}{c|}{4.295} & 45.71 & \multicolumn{1}{c|}{85} & 86 & -1                    \\ \hline
minichat-3b                  & \multicolumn{1}{c|}{4.211} & 48.82 & \multicolumn{1}{c|}{86} & 83 & 3                     \\ \hline
pythia-12b-mix-sft           & \multicolumn{1}{c|}{4.173} & 41.86 & \multicolumn{1}{c|}{87} & 88 & -1                    \\ \hline
alpaca-7b                    & \multicolumn{1}{c|}{4.171} & 26.46 & \multicolumn{1}{c|}{88} & 91 & -3                    \\ \hline
chatglm2-6b                  & \multicolumn{1}{c|}{4.169} & 47.13 & \multicolumn{1}{c|}{89} & 84 & 5                     \\ \hline
phi-2                        & \multicolumn{1}{c|}{4.164} & 30.66 & \multicolumn{1}{c|}{90} & 90 & 0                     \\ \hline
oasst-sft-pythia-12b         & \multicolumn{1}{c|}{4.087} & 25.96 & \multicolumn{1}{c|}{91} & 92 & -1                    \\ \hline
text\_davinci\_001           & \multicolumn{1}{c|}{3.927} & 15.17 & \multicolumn{1}{c|}{92} & 95 & -3                    \\ \hline
falcon-7b-instruct           & \multicolumn{1}{c|}{3.919} & 23.60 & \multicolumn{1}{c|}{93} & 93 & 0                     \\ \hline
text\_davinci\_003           & \multicolumn{1}{c|}{3.891} & 50.00 & \multicolumn{1}{c|}{94} & 82 & 12                    \\ \hline
baichuan-13b-chat            & \multicolumn{1}{c|}{3.787} & 21.80 & \multicolumn{1}{c|}{95} & 94 & 1                     \\ \hline

\end{longtable}

\end{document}